\documentclass[journal,hideappendix]{vgtc}        

\DeclareGraphicsExtensions{.png,PNG,.pdf,.PDF}

\onlineid{1405}

\vgtccategory{Research}

\def \sys {{Uni-Evaluator}}

\def \etal {{\emph{et al}.\thinspace}}

\def \etc {{\emph{etc}.\thinspace}}
\def \eg {{\emph{e.g}.\thinspace}}

\title{A Unified Interactive Model Evaluation for Classification, Object Detection, and Instance Segmentation in Computer Vision}

\author{%
  Changjian Chen,
  Yukai Guo,
  Fengyuan Tian,
  Shilong Liu,
  Weikai Yang, \\
  Zhaowei Wang,
  Jing Wu,
  Hang Su,
  Hanspeter Pfister,
  and Shixia Liu
}

\authorfooter{
  \item
  	C.~Chen, Y.~Guo, F.~Tian, W.~Yang, Z.~Wang, and S.~Liu are with the School of Software, BNRist, Tsinghua University. C.~Chen and Y.~Guo are joint first authors. S.~Liu is the corresponding author.
  	E-mail: \{\{ccj17, gyj22, tianfy21, yangwk21, wzw20\}@mails., shixia@\}tsinghua.edu.cn.
   \item
        S.~Liu, and H.~Su are with the Department of Computer Science and Technology, Tsinghua University.
        E-mail: slongliu86@gmail.com, suhangss@tsinghua.edu.cn.
   \item
        J.~Wu is with Cardiff University.
        E-mail: wuj11@cardiff.ac.uk.
  \item
  	H.~Pfister is with Harvard University.
  	E-mail: pfister@seas.harvard.edu.
}

\abstract{Existing model evaluation tools mainly focus on evaluating classification models, leaving a gap in evaluating more complex models, such as object detection.
In this paper, we develop an open-source visual analysis tool, {\sys}, to support a unified model evaluation for classification, object detection, and instance segmentation in computer vision.
The key idea behind our method is to formulate both discrete and continuous predictions in different tasks as unified probability distributions.
Based on these distributions, we develop 1) a matrix-based visualization to provide an overview of model performance; 2) a table visualization to identify the problematic data subsets where the model performs poorly; 3) a grid visualization to display the samples of interest.
These visualizations work together to facilitate the model evaluation from a global overview to individual samples.
Two case studies demonstrate the effectiveness of {\sys} in evaluating model performance and making informed improvements.}

\keywords{Model evaluation, computer vision, classification, object detection, instance segmentation}

\teaser{
  \centering
  \includegraphics[width=\linewidth]{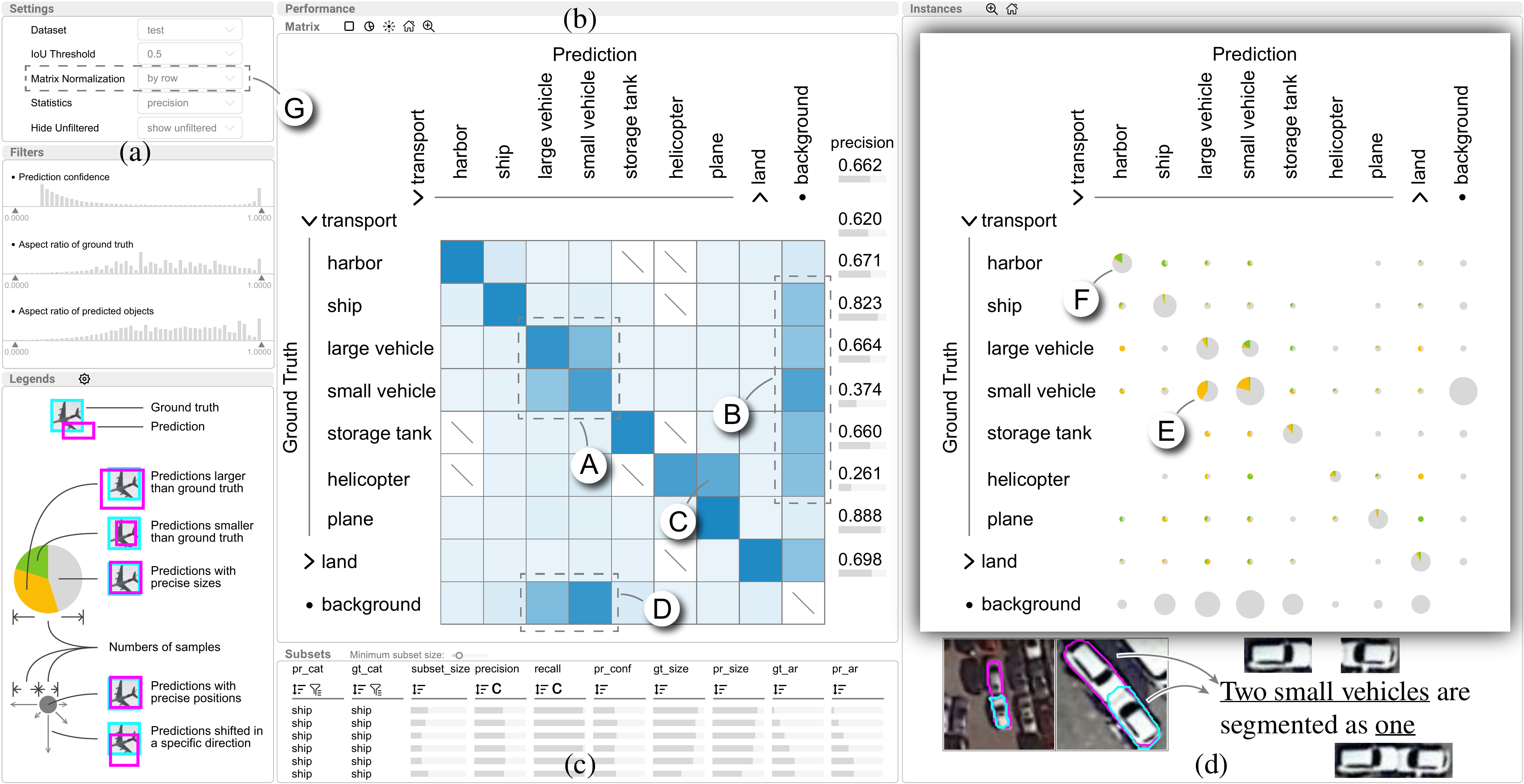}
  \caption{{\sys}: 
  (a) the filtering panel;
  (b) the matrix-based visualization provides an overview of model performance;
  (c) the table visualization helps identify problematic data subsets;
  (d) the grid visualization displays the samples of interest.}
  \label{fig:teaser}
}

\graphicspath{{figs/}{figures/}{pictures/}{images/}{./}} 

\usepackage{multirow}
\usepackage{color}
\usepackage{lipsum}                    
\usepackage{amssymb,amsmath}
\usepackage{wrapfig}
\usepackage{booktabs}
\usepackage{makecell}
\usepackage{dblfloatfix}
\usepackage{pdfpages}

\newcommand{\myparagraph}[1]{\vspace{1mm}\noindent\textbf{#1}}

\newcommand{\glyph}[2][1.3]{$\vcenter{\hbox{\includegraphics[height=#1\fontcharht\font`\B]{figures/glyph/#2.pdf}}}$}

\usepackage{mathptmx}                  

\begin{document}


\firstsection{Introduction}

\maketitle
\fontsize{9}{9} 
Model evaluation assesses performance of machine learning models and helps identify the causes of poor performance for further improvement~\cite{ren2016squares}.
It is a critical step in the development of machine learning models.
For example, machine learning practitioners usually evaluate models generated with different parameters and select the best one for deployment.
Currently, the most widely-used way for evaluating models is to use performance metrics, such as accuracy and mAP (mean Average Precision).
However, using these metrics alone can sometimes be misleading. 
For example, a classification model that always predicts the majority class in an imbalanced dataset can achieve high accuracy but has no predictive power at all~\cite{johnson2019survey}. 
To better understand model performance, evaluation at finer granularity is required. 
Confusion matrices~\cite{townsend1971theoretical} are thus developed, which provide evaluations at the class level by presenting confusion between classes.
However, they only focus on the classification task where the accuracy of classifying samples is of concern.
In addition to classification, there are other tasks that are popularly used in real-world applications~\cite{liu2020deep, minaee2021image}. 
Their model evaluations concern more than the accuracy of sample-level classification. 
For example, object detection concerns not only the accuracy of classifying objects in images, but also the precision of locating the objects, which is measured by mAP.
For these tasks, confusion matrices are not directly applicable.
In addition, like object detection, there are scenarios where multiple tasks are considered simultaneously~\cite{zhang2021survey}.
Although we can develop different fine-grained 
model evaluation methods for different tasks, it will increase the learning cost and cognitive load of users when switching among them~\cite{schroeder2008switching}.
Thus, a unified model evaluation method for different tasks in computer vision is desirable.

From a review of previous studies and a survey with 151 computer vision experts, 
we identified two key challenges in developing a unified model evaluation method.
First, predictions in different tasks can be discrete (\eg, class labels) or continuous (\eg, positions of objects).
To analyze these predictions in a single tool, we need to model them in a unified way.
While previous research has explored how to model different discrete predictions~\cite{gortler2022neo}, 
it is largely unexplored how to jointly model discrete and continuous predictions.
Second, users not only evaluate overall performance on the entire dataset, but are also interested in analyzing performance at different levels of detail, such as subsets and individual samples.
For example, users are interested in problematic subsets that are sliced along different attributes, such as class labels or object sizes.
Examples of such subsets include the samples of a class with low accuracy in a classification task and the small objects that the model failed to detect in an object detection task.
Finding these problematic subsets in a large dataset and examining their predictions are difficult and time-consuming.
Therefore, a multi-level exploration environment for efficient model evaluation is necessary.

To tackle these challenges, we develop {\sys}, a unified interactive model evaluation method for computer vision tasks.
We focus on three main tasks in computer vision: classification, object detection, and instance segmentation~\cite{bolya2020tide}.
To support a unified analysis of model performance in different tasks, we propose a unified formulation that extends the discrete probability formulation proposed by G{\"o}rtler~\etal~\cite{gortler2022neo}.
The key idea is to formulate both discrete and continuous predictions as unified probability distributions.
Based on these distributions, we first develop a matrix-based visualization with three evaluation modes.
It provides an overview of model performance.
Next, a table visualization is developed based on LineUp~\cite{gratzl2013lineup}.
It is supported by a frequent pattern mining-based search method to facilitate the identification of problematic data subsets.
In addition, a grid visualization is used to display the samples of interest.
With the three coordinated visualizations, users can comprehensively evaluate model performance from a global overview to individual samples and make informed improvements.
{\sys} relies only on the input and output of models to perform the evaluation, so it is model-agnostic.
Two case studies are conducted on object detection and instance segmentation, respectively, to validate the contributions of {\sys}.
A Python package is released at \href{http://uni-evaluator.thuvis.org/}{http://uni-evaluator.thuvis.org/}.

The contributions of our work include:

\begin{itemize}[nosep]
\item\noindent{\textbf{Coordinated visualizations} to explain model performance at different levels, which include a matrix-based visualization with three evaluation modes, a table visualization supported by frequent pattern mining-based search, and a grid visualization.} \looseness=-1
\item\noindent{\textbf{A unified probability distribution method} that jointly models discrete and continuous predictions for evaluating different models in one tool.}
\item\noindent{\textbf{An open-source visual analysis tool} that integrates the probability distribution with the coordinated visualizations to support a unified model evaluation for computer vision tasks.}

\end{itemize}

\section{Related Work}

\subsection{Model Evaluation in Computer Vision}
In the field of computer vision, the most widely-used method for evaluating models is to use performance metrics such as accuracy, mAP, and mIoU.
Some recent methods further classify model errors into multiple types for understanding model performance~\cite{hoiem2012diagnosing, borji2019empirical, bolya2020tide}.
For example, Bolya~\etal~\cite{bolya2020tide} proposed to categorize object detection errors into six disentangled ones to provide a detailed performance summary from different perspectives.
Although these methods are useful in quantitatively summarizing model performance, 
they fail to identify the causes of poor performance.
Compared with these methods, our work combines a matrix-based visualization, a table visualization, and a grid visualization to 
evaluate model performance from high-level metrics to individual samples.
Such a comprehensive evaluation enables users to diagnose poor performance of a computer vision model and make informed improvements to the model and/or the associated data.

\subsection{Model Evaluation in Visualization}
In the field of visualization, model evaluation methods are classified into two categories: model-specific and model-agnostic methods.
Model-specific methods analyze the inner components of a specific type of models, such as the convolutional layers of deep neural networks~\cite{hohman2018visual, liu2016towards, liu2018analyzing, yuan2021survey, lei2020geometric}.
Model-agnostic methods explain model performance by considering the inputs and outputs of models regardless of their model types.
Our work falls into the latter category, 
where the methods are divided into three groups: class level, instance level, and their combination.\looseness=-1

Class-level evaluations focus on analyzing the confusion between classes.
The most widely-used method is the confusion matrix~\cite{townsend1971theoretical}.
This matrix shows which classes are confused with each other and how heavy the confusion is.
Later efforts focus on generalizing the confusion matrix to other scenarios~\cite{gleicher2020boxer, gortler2022neo, xenopoulos2022calibrate, hinterreiter2020confusionflow, heyen2020clavis}.
For example, Boxer~\cite{gleicher2020boxer} compares performance of different classifiers on the selected subsets by combining a confusion matrix with several chart visualizations.
ConfusionFlow~\cite{hinterreiter2020confusionflow} enhances
the confusion matrix with stacked heatmaps and line charts to support the analysis of class confusion over time.\looseness=-1

Instance-level evaluations seek to analyze the predictions of individual samples and their similarity relationships.
Scatterplots and grid visualizations are the two most widely-used instance-level methods.
In a scatterplot, each sample is represented by a dot with the color encoding its predicted or ground-truth class label.
The positions of the dots are determined by their sample attributes (\eg, prediction confidence~\cite{amershi2015modeltracker}),
or projection (\eg, t-SNE~\cite{maaten2008visualizing, li2022unified, yang2020drift}) of their sample attributes.
Grid visualizations place the content of samples (\eg, images) in grids, which overcomes the overlapping and space-wasting issues of scatterplots.
Along this line, Chen~\etal~\cite{chen2020oodanalyzer} presented a hierarchical grid visualization to support the exploration of a large number of samples.
To maintain mental maps during exploration, DendroMap~\cite{bertucci2022dendromap} 
enhances the grid visualization with an interactive zoomable treemap.

In real-world settings, both class-level and instance-level evaluations are required to understand model performance. 
Accordingly, these two evaluation methods are combined to maximize the values of both.
The most intuitive combination is to use separate but coordinated views~\cite{alsallakh2014visual, bilal2017convolutional, chen2021interactive, zhang2022sliceteller, abadi2016tensorflow, biewald2020experiment}.
For example, ConfusionWheel~\cite{alsallakh2014visual} utilizes a chord diagram augmented with histograms to display confusion between classes and the prediction confidence distribution of each class.
Users can inspect a subset of selected samples in scatterplots and examine their feature distribution in bar charts.
Such a strategy is also utilized by TensorBoard~\cite{abadi2016tensorflow} and Weights\&Biases~\cite{biewald2020experiment}.
Although these methods enable the simultaneous examination of the class-level metrics and instance-level predictions, users have to switch views between the two levels. 
To tackle this issue, Squares~\cite{ren2016squares} combines the two-level information using stacked bar charts,
which utilize bars to display confusion between classes and small squares in each bar to represent the associated samples.
\looseness=-1

Despite their effectiveness,
these methods only consider single-output labels and fail to support more complex data structures, such as multi-output labels.
To solve this problem, a pioneering work, Neo~\cite{gortler2022neo}, employs probability distributions to represent confusion matrices. 
These distributions generalize the traditional confusion matrix to support hierarchical and multi-output labels.
However, Neo primarily focuses on classification tasks and lacks support for object detection and instance segmentation tasks.
To provide a unified model evaluation method for different computer vision tasks, we extend the discrete formulation in Neo to jointly model discrete and continuous predictions presented in different tasks.
Moreover, we develop three coordinated visualizations to facilitate the identification of factors contributing to poor performance.\looseness=-1

\section{Survey-Based Task Analysis and System Design}

\begin{figure}[b]
\vspace{-3mm}
    \centering
    \includegraphics[width=\linewidth]{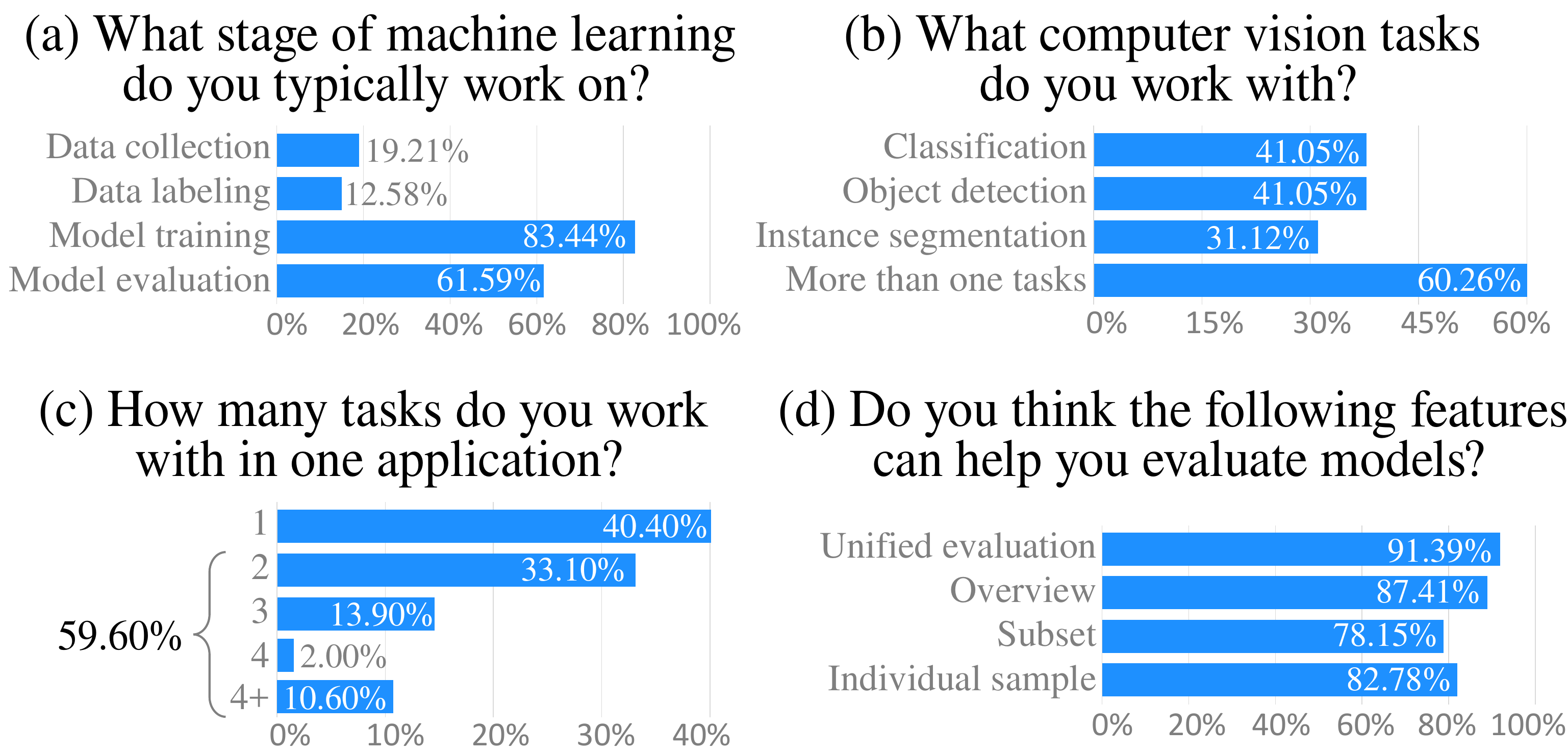}
\vspace{-3mm}
    \caption{Survey responses: 
    (a) the participants mainly focused on model training and evaluation; 
    (b) 91 participants ($60.26\%$) had worked with more than one task;
    (c) 90 participants ($59.60\%$) had worked with multi-task applications;
    (d) most participants thought unified evaluation and multi-level exploration are (very) important.
    }
    \label{fig:survey}
\end{figure}

\begin{figure*}[t]
    \centering
    \includegraphics[width=\linewidth]{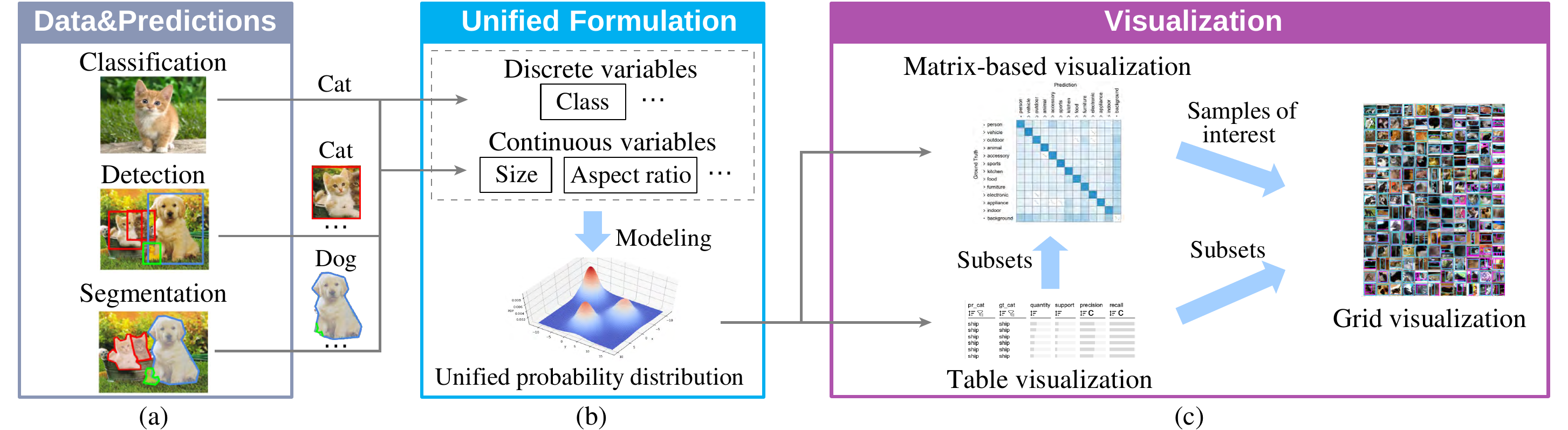}
    \vspace{-5mm}
    \caption{System overview: (a) given data samples and model predictions with both discrete and continuous variables, (b) the unified formulation module models them by a probability distribution; (c) the visualization module explains model performance at different levels.}
    \label{fig:pipeline}
    \vspace{-5mm}
\end{figure*}

To better understand the current practice of model evaluation and the challenges computer vision experts face, we conducted an expert survey. 
We first designed a survey based on the literature review.
Then we discussed it with four computer vision experts ($E_1$--$E_4$) for refinement.
The survey covered three parts: 1) the basic information of the participants (Q1--Q3); 2) current practices in evaluating computer vision models (Q4--Q8); 3) key features needed for a unified model evaluation (Q9--Q12).
The exact questions can be found in supplemental material.

We distributed the refined survey to the computer vision experts from three top universities and three major technology companies.
A total of 151 surveys were returned.
Among the participants, 28 ($18.54\%$) had less than 1 year of experience in computer vision, 50 ($33.11\%$) had 1-3 years of experience, 50 ($33.11\%$) had 3-5 years of experience, 23 ($15.23\%$) had more than 5 years of experience.
In addition, the participants spent more attention on ``model training'' ($83.44\%$) and ``model evaluation'' ($61.59\%$) but less attention on ``data collection and processing'' ($19.21\%$) and ``data labeling'' ($12.58\%$) (Fig.~\ref{fig:survey}(a)). 
This indicates that they have extensive experience to speak about the intricacies of model evaluation.

\subsection{Why Unified Evaluation}
\label{subsec:why-unified}
The survey results indicated that most of the participants ($91.39\%$) were interested in a unified model evaluation tool (Fig.~\ref{fig:survey}(d)).
From their responses, we identified three main reasons.

\myparagraph{Reducing the learning curve}.
One major advantage of a unified model evaluation tool is that it can reduce the time for users to learn different tools for different tasks.
According to the survey results, 91 participants ($60.26\%$) had worked on more than one task (Fig.~\ref{fig:survey}(b)).
These participants pointed out that they had to use different tools to evaluate different tasks.
This significantly increased the time they spent on learning the different visual encodings and functionalities of these tools.
Moreover, different tools require different formats of data input.
Users have to maintain different data pre-processing codes, which is also a burden for them.

\myparagraph{Improving the analysis efficiency}.
As shown in Fig.~\ref{fig:survey}(c), 90 participants ($59.60\%$) had experience in multi-task applications.
However, when evaluating application models, users still need to use multiple tools for multiple tasks. 
Their analysis is often interrupted as they have to switch between different tools.
For example, a participant reported that while developing models for classifying medical images, he often carried out the segmentation of lesions.
When performance is poor, he uses a confusion matrix to identify the confused classes.
Then, to identify the main causes of the confusion, 
he filters the associated samples and visualizes their segmentation results in a different tool. This significantly reduces the analysis efficiency.

\myparagraph{Debugging a multi-task model}.
In addition to reducing the learning curve and improving the analysis efficiency, 
a unified model evaluation will also help debug a multi-task model effectively.
In a multi-task application, the tasks mutually influence each other. 
Understanding such mutual influence helps users debug the model~\cite{zhang2021survey, zhang2018overview}, such as identifying which task is the weak point of the model.
Currently, different tasks are analyzed with different tools, which hinders the effective analysis of the mutual influence between tasks.
Therefore, a unified evaluation for multiple tasks is desirable.

\subsection{Design Goals}
To distill the detailed design goals and tasks for developing such a unified tool, we further conducted interviews with four experts ($E_1$--$E_4$) for more detailed feedback and insights.

\myparagraph{G1 - Evaluating different models with a unified tool}.
According to the survey results, 138 participants (91.39\%) wanted a unified tool for evaluating different models, as switching between multiple tools is inconvenient.
The main obstacle to building a unified tool is that the predictions in different tasks can be discrete or continuous.
Although Neo~\cite{gortler2022neo} has explored the modeling of discrete predictions, it is still under exploration how to model both discrete and continuous predictions.\looseness=-1

\myparagraph{G2 - Analyzing model performance at different levels}.
The participants also expressed their need to analyze model performance at different levels.
As shown in Fig.~\ref{fig:survey}(d), most participants were interested in analyzing model performance at dataset level, subset level, and sample level.
In current practice, they often utilize performance metrics or confusion matrices to get an overview of model performance.
However, these methods may hinder the identification of problematic subsets where a model performs poorly.
Analyzing model performance on such subsets helps improve the robustness of the model.
In addition, when diagnosing poor performance on these subsets,
Therefore, a multi-level exploration environment is needed.\looseness=-1

\myparagraph{G3 - Finding problematic data subsets}. 
As mentioned above, the participants needed to analyze model performance on problematic subsets.
However, identifying such problematic subsets is non-trivial, especially when the dataset is large.
Currently, they usually find such subsets by manually creating different rules to slice the data and then examining these subsets one by one.
This process is tedious and time-consuming.
Therefore, the participants wanted a more efficient way to identify the problematic subsets.

\subsection{Task Analysis}
Based on the design goals, we derived several tasks as guidelines for designing the unified model evaluation tool.

\myparagraph{T1 - Modeling both discrete and continuous predictions in a unified manner (\textbf{G1})}.
This includes a unified formulation for both discrete and continuous predictions to enable the analysis of model performance across different tasks.

\myparagraph{T2 - Visually explaining overall performance on the entire dataset (\textbf{G2})}.
This includes the confusion between classes and imprecise sizes and positions of predicted objects.

\myparagraph{T3 - Interactively identifying problematic data subsets (\textbf{G2}, \textbf{G3})}.
This includes a subset search method to extract candidate subsets and a table visualization that allows users to identify problematic data subsets based on one or more attributes.

\myparagraph{T4 - Displaying the samples of interest for efficient exploration (\textbf{G2})}.
This includes a grid visualization that places similar samples together and visually presents them in a compact way to facilitate exploration.

\subsection{Design of {\sys}}
Motivated by the identified tasks, we develop {\sys} to support a unified model evaluation for different computer vision tasks.
As shown in Fig.~\ref{fig:pipeline}, {\sys} consists of two main modules: \textbf{unified formulation} (Sec.~\ref{sec:formulation}) and \textbf{visualization} (Sec.~\ref{sec:visualization}).

Given data samples and model predictions with both discrete and continuous variables (Fig.~\ref{fig:pipeline}(a)),
the unified formulation module models them by a probability distribution (\textbf{T1}, Fig.~\ref{fig:pipeline}(b)).
Based on the distribution, the visualization module provides three coordinated visualizations to explain model performance at different levels (Fig.~\ref{fig:pipeline}(c)).
The matrix-based visualization provides an overview of model performance (\textbf{T2}).
Candidate data subsets are extracted and presented in the table visualization.
Users can identify the problematic subsets (\textbf{T3}) and analyze their predictions in the matrix-based visualization.
In both visualizations, users can select the samples of interest and explore them in the grid visualization (\textbf{T4}).
With {\sys},
users can diagnose the causes of poor performance and make informed improvements.

\begin{figure}[!b]
\vspace{-5mm}
    \centering
    \includegraphics[width=\linewidth]{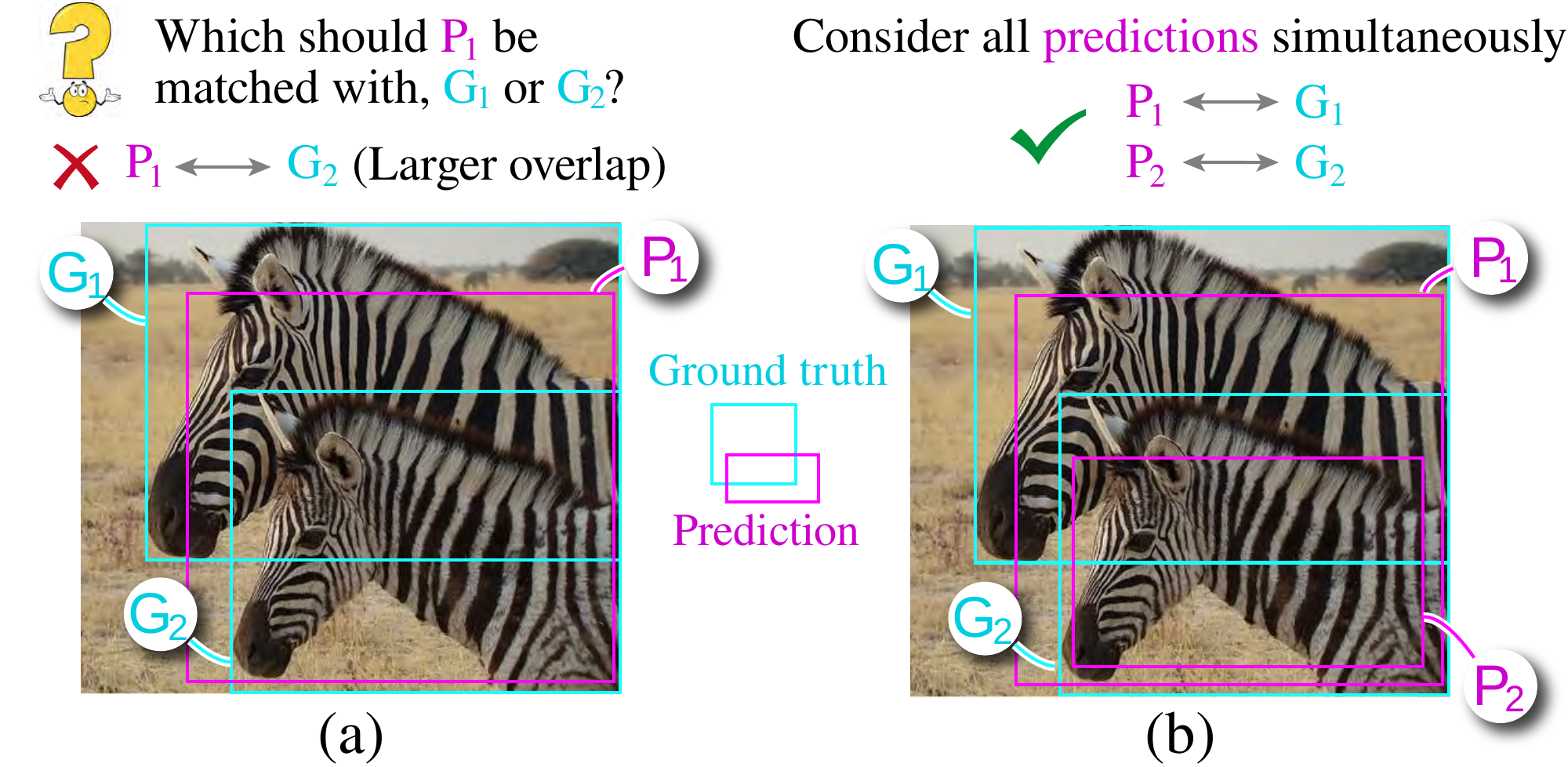}
    \vspace{-5mm}
    \caption{In object matching, 
    (a) treating each prediction independently results in incorrect matches, while (b) considering all predictions simultaneously enables the inference of the optimal matching.}
    \label{fig:match}
\end{figure}

\section{Unified Probability Distribution}
\label{sec:formulation}

Neo~\cite{gortler2022neo} utilizes probability distributions to model discrete variables in classification tasks by matching predicted classes with their ground truth.
We extend it to also model continuous predictions and ground truth.
However, in this probabilistic framing, it is non-trivial to match each continuous prediction with the corresponding ground truth due to multiple possible ground-truth annotations.
To tackle this issue, we develop an object matching algorithm to pair each prediction with the suitable ground truth.
Based on the matching results, we model discrete and continuous variables using a unified probability distribution.
\looseness=-1

\myparagraph{Object matching}.
A straightforward way to match a prediction with its ground truth is to find the one with the largest overlap.
However, this strategy considers each prediction separately and may not capture the optimal matches between predictions and their associated ground truth.
For example, in Fig.~\ref{fig:match}(a), the detected object P${}_1$ has a larger overlap with the ground truth G${}_2$.
By only considering P${}_1$,
it will be matched with G${}_2$.
However, 
if another detected object, P${}_2$, is considered simultaneously (Fig.~\ref{fig:match}(b)),
the matching between P${}_1$ and G${}_2$ is incorrect.
The optimal way is to match P${}_1$ with G${}_1$ and P${}_2$ with G${}_2$.
Therefore, it needs to consider all the predictions for optimal matching.
Mathematically, the matching between the predictions and ground truth can be modeled as a one-to-many assignment problem~\cite{pentico2007assignment},
which maximizes the total assignment award. \looseness=-1
\begin{equation}
\begin{aligned}
\underset{z_{ij}}{\mathrm{maximize}}\quad&\textstyle\sum_{j}^{M_p}\sum_{i}^{M_g} z_{ij}a_{ij}\\
\mathrm{s.t.}\quad z_{ij}\in\{0,1\},\quad z_{ij}\mathbb{I}&(p_{ij}<\alpha)=0,\quad\textstyle\sum_i z_{ij}\leq 1,\ \forall i,j.
\label{eq:match}
\end{aligned}
\end{equation}
$M_p$ and $M_g$ are the numbers of predictions and ground truth, respectively.
$z_{ij}$ is a binary variable indicating whether the $i$-th ground truth is matched with the $j$-th prediction.
$a_{ij}$ is the award for matching the $i$-th ground truth with the $j$-th prediction.
A larger value of $a_{ij}$ implies a stronger matching between the two.
It consists of three parts.
\begin{itemize}[nosep]
    \item \noindent
    \emph{Label consistency score} ($c_{ij}$) that ensures the
    matched prediction and ground truth have the same label.
    It is set to 1 when the prediction and ground truth have the same label, and 0 otherwise. 
    \item \noindent
    \emph{Position consistency score} ($p_{ij}$) that ensures the
    matched prediction and ground truth have a large overlap.
    We apply the widely-used IoU (Intersection over Union) to measure the overlap.\looseness=-1
    \item \noindent
    \emph{Uniqueness score} of the $i$-th ground truth ($u_{i}$) that ensures the ground truth is matched with one prediction.
    We set $u_{i}=e^{-\sum_{k}z_{ik}}$.
    The more predictions that are matched with the $i$-th ground truth, the smaller $u_{i}$.
\end{itemize}
$a_{ij}$ is defined as the weighted sum of the three scores: $a_{ij} = \lambda_1 c_{ij} +\\ \lambda_2 p_{ij} + (1-\lambda_1-\lambda_2) u_{i}$.
$\lambda_1$ and $\lambda_2$ are used to balance the three parts.
They are set as $0.5$ and $0.25$
to emphasize the importance of the label consistency score~\cite{bolya2020tide}.
The constraint $z_{ij}\mathbb{I}(p_{ij}<\alpha)=0$ requires that the matched prediction and ground truth should have a minimum overlap $\alpha$.
$\alpha$ is set to $0.1$ following the setting in~\cite{bolya2020tide}.
$\mathbb{I}(\cdot)$ is the indicator function.
Constraint $\sum_i z_{ij}\leq 1$ ensures that each prediction is matched to a maximum of one ground truth.
We solve this assignment problem with a greedy strategy.
We match the predictions in an image one by one according to their confidence in descending order. 
For each prediction, we match it with a ground truth to maximize Eq.~(\ref{eq:match}). 
Thus, the time complexity to match the predictions with the ground truth in an image is $O(M_pM_g)$, and for a dataset with $N$ images, it is $O(M_pM_gN)$.
As $M_p$ and $M_g$ are usually significantly smaller than $N$ in a large dataset, the time complexity is almost $O(N)$. 
For example, it only takes four minutes to process the COCO dataset~\cite{lin2014microsoft} with more than 100,000 images. 
In addition, object matching is a one-time pre-processing step. 
Thus, this method can effectively scale up to handle large datasets with millions of images.
\looseness=-1

\myparagraph{Modeling both discrete and continuous variables}.
Based on the matching
between predictions and ground truth,
we adopt the joint probability distribution to model both discrete and continuous variables.
Let $C$ and $D$ denote continuous and discrete variables, and let $X$ and $Y$ denote ground truth and predictions.
Then the joint probability distribution is given by $P(C_X, D_X, C_Y, D_Y)$.
In \sys, discrete variables include predicted/ground-truth classes,
and continuous variables include the sizes and aspect ratios of predicted objects and ground truth, and prediction confidence.

To determine the corresponding probability function,
we need to consider the probability of both discrete and continuous variables.
For discrete variables,
the probability mass function is used to determine the probability of taking on a specific value, which is determined by the frequency of the variable equal to that value. 
For continuous variables,
the probability is calculated over intervals using the cumulative distribution function (CDF), which is determined by the frequency of the variable in that interval. 
In our implementation, we utilize the empirical CDF because of its time efficiency~\cite{jadhav2009parametric}.
This method discretizes a continuous variable and counts the frequency of the discretized variable less than a specific value.
With the two functions,
the joint probability is calculated using the conditional probability.
Here, we illustrate the calculation of the joint probability using an object detection example. 
Suppose that we want to analyze the detected objects with the ground-truth label ``cat'' and high prediction confidence ($>0.5$).
This is defined as
$P(\mathrm{Label}_X=\mathrm{cat},\ \mathrm{Confidence}_Y>0.5)$.
The probability can then be calculated using
the following conditional probability:
\begin{equation}
P(\mathrm{Confidence}_Y>0.5\mid\mathrm{Label}_X=\mathrm{cat})P(\mathrm{Label}_X=\mathrm{cat}),
\end{equation}
where $P(\mathrm{Label}_X=\mathrm{cat})$ is determined by the probability mass function of discrete variable $\mathrm{Label}_X$, and 
$P(\mathrm{Confidence}_Y>0.5\mid\mathrm{Label}_X=\mathrm{cat})$ is determined
by the cumulative distribution function of continuous variable $\mathrm{Confidence}_Y$ given $\mathrm{Label}_X$.

The probabilistic framework enables users to process data for different analysis tasks using the standard operations of probability distributions,
including marginalization and conditioning~\cite{gortler2022neo}.
The \textbf{marginalization} operation discards variables in the distribution by integrating or summing them,
allowing users to focus on analyzing the variables of interest.
The \textbf{conditioning} operation constrains variables to be of specific values or within specific intervals,
which enables users to analyze the subsets of interest.

\section{{\sys} Visualization}
\label{sec:visualization}
Based on the unified probability distribution, we developed three coordinated visualizations to explain model performance at different levels: 1) a matrix-based visualization (Fig.~\ref{fig:teaser}(b)) to provide an overview of model performance (\textbf{T2}); 2) a table visualization (Fig.~\ref{fig:teaser}(c)) to identify problematic subsets (\textbf{T3}); and 3) a grid visualization (Fig.~\ref{fig:teaser}(d)) to display the samples of interest (\textbf{T4}).

\subsection{Matrix-based Visualization}
To evaluate both classification and localization performance in a unified manner, it is essential to consider 1) the confusion between classes; 
2) the sizes of predicted objects; and 3) the directions in which they are shifted from ground truth.
However, analyzing the three interdependent aspects simultaneously can be challenging.
Therefore, for more effective analysis, we disentangle them by utilizing the marginalization operation to discard irrelevant variables.
Accordingly, three evaluation modes are provided:
1) confusion mode for evaluating classification performance; 2) size mode for analyzing the sizes of predicted objects; 3) direction mode for analyzing the shifted directions of predicted objects.
Within each mode, we disentangle the associated variables from others by utilizing the marginalization operation to discard irrelevant variables.
\looseness=-1

We employ a matrix-based visualization, an extension of Neo~\cite{gortler2022neo}, to convey the information in the three modes. Specifically, the size/direction mode is enhanced with a carefully designed glyph to support the identification and diagnosis of size/direction errors.
The rows of the matrix represent ground-truth classes, and the columns represent predicted classes.
The class names are displayed left to the matrix and above the matrix. 
When the number of classes is large (\eg, 30 in a view with a resolution of $800\times800$), the matrix cells become small, which reduces their visibility.
To address this, we organized the classes hierarchically using hierarchical clustering algorithms~\cite{yang2021interactive, yang2022diagnosing, murtagh2012algorithms} or based on their inherent hierarchical structure, and presented them as an indented tree.
Furthermore, users can enlarge the cells of interest with the drill-down and hovering interactions, which are described at the end of this section.
The summary statistics of each class (precision, recall,~\etc) is displayed as a list on the right side of the matrix.
The only difference between the three modes is the content within the matrix cells.
Next, we delve into each of these modes.
\looseness=-1

\begin{wrapfigure}{l}{0.06\textwidth}
\vspace{-12pt}
\includegraphics[width=0.06\textwidth]{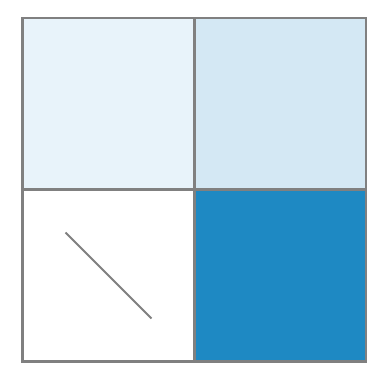}
\vspace{-14pt}
\end{wrapfigure}
\noindent \emph{Confusion mode}.
In confusion mode, the cell represents the number of samples in classification tasks and the number of objects in detection/segmentation tasks. 
The frequency of the samples/objects confused between classes is encoded by the filling color in each cell, ranging from white to blue.
The darker the cell color, the greater the confusion between the associated classes. 
Same to Neo~\cite{gortler2022neo}, we place a light-gray dash in the cells with no samples/objects to make them more distinguishable from those with only a few samples/objects.
To make the confusion patterns more distinct, we employ the Optimal Leaf Ordering algorithm~\cite{bar2001fast} to reorder the matrix.
\looseness=-1

\begin{wrapfigure}{l}{0.06\textwidth}
\vspace{-12pt}
\includegraphics[width=0.06\textwidth]{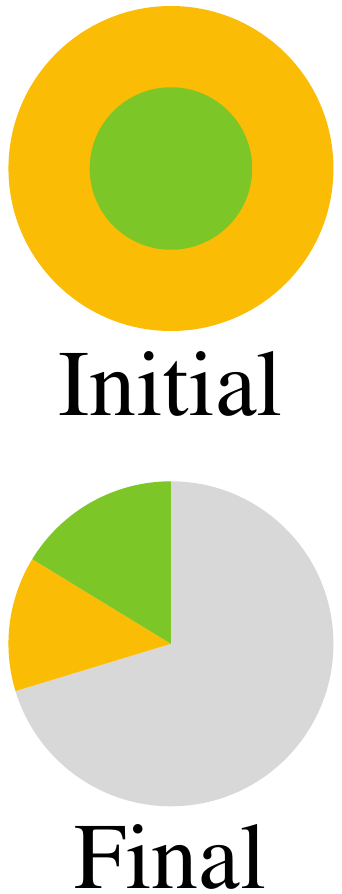}
\vspace{-14pt}
\end{wrapfigure}
\noindent \emph{Size mode}.
Initially, we utilize two concentric circles to represent the size error.
The green circle represents the predicted objects with sizes smaller than the ground truth, and the yellow circle represents the predicted objects with larger sizes.
Their radii encode the number of associated objects.
The experts agree that this design is intuitive in identifying the main size errors in the cells.
However, they are also interested to know how many predicted objects are with precise sizes, which is not shown in the initial design.
To address this, we utilize a pie chart with three sectors to summarize the sizes of predicted objects in each cell.
The gray sector represents predicted objects with precise sizes, while the yellow/green represents those with larger/smaller sizes compared to ground truth.
To address the concern of color blindness, we provide the option for users to customize the color encoding themselves by clicking \glyph[1]{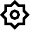}.
The angle of each sector is proportional to the number of predicted objects in this sector,
and the radius of the pie chart encodes the number of predicted objects in that cell.
\looseness=-1

\begin{wrapfigure}{l}{0.06\textwidth}
\vspace{-12pt}
\includegraphics[width=0.06\textwidth]{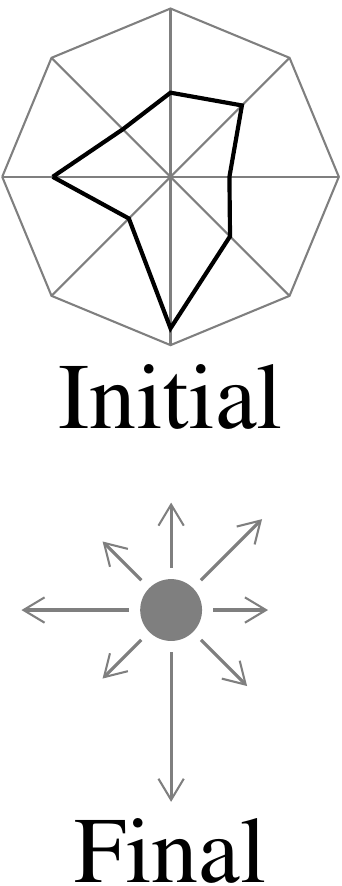}
\vspace{-14pt}
\end{wrapfigure}
\noindent \emph{Direction mode}.
A straightforward way for encoding the shifted directions of predicted objects is to utilize a radar chart with eight spokes.
The length of each spoke encodes the number of predicted objects shifted in that direction.
A polygon is drawn to connect the spokes.
When showing this design to the experts, 
they indicate that the polygon shapes immediately attract their attention.
However, the shapes provide less information in understanding the predictions compared with the lengths of the spokes. 
Additionally, redundant lines in the radar chart, such as the boundary lines and the extended lines along the spokes, make the visual representation even more complex
and add extra cognitive load to users. 
To address this issue, we replace the radar chart with eight arrows.
The length of each arrow encodes the number of predicted objects shifted in that direction.
We also add a circle in the middle to represent the predicted objects with precise positions.
Its radius encodes the number of such objects.\looseness=-1

In the matrix-based visualization, several interactions are provided to facilitate exploration.
First, users can use the conditioning operation to drill down into sub-matrices of interest by clicking \glyph[1]{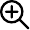} at the top of Fig~\ref{fig:teaser}(b).
In addition, when a class has much more samples/objects than the others, some patterns in the matrix may be hidden.
For example, in the confusion mode, 
the cell colors of the dominant classes can overshadow other classes (\eg, \glyph[1.5]{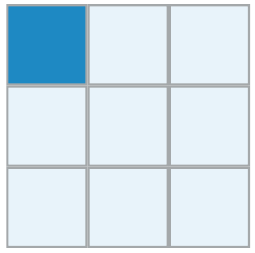}), making it difficult to detect misclassifications in the other classes.
To address this issue, row/column normalization is supported (Fig.~\ref{fig:teaser}G).
For example, by performing row normalization on the matrix~\glyph[1.5]{matrix}, the confusion between the second and third classes appears (\glyph[1.5]{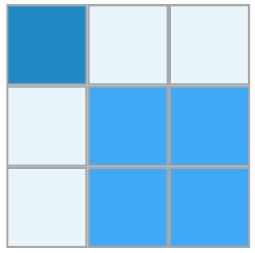}).
Moreover, in the size and the direction modes, the charts with few objects are usually small.
To ensure their visibility when needed, users can enlarge them upon hovering.

\begin{figure*}[!b]
    \vspace{-5mm}
    \centering
    \includegraphics[width=\linewidth]{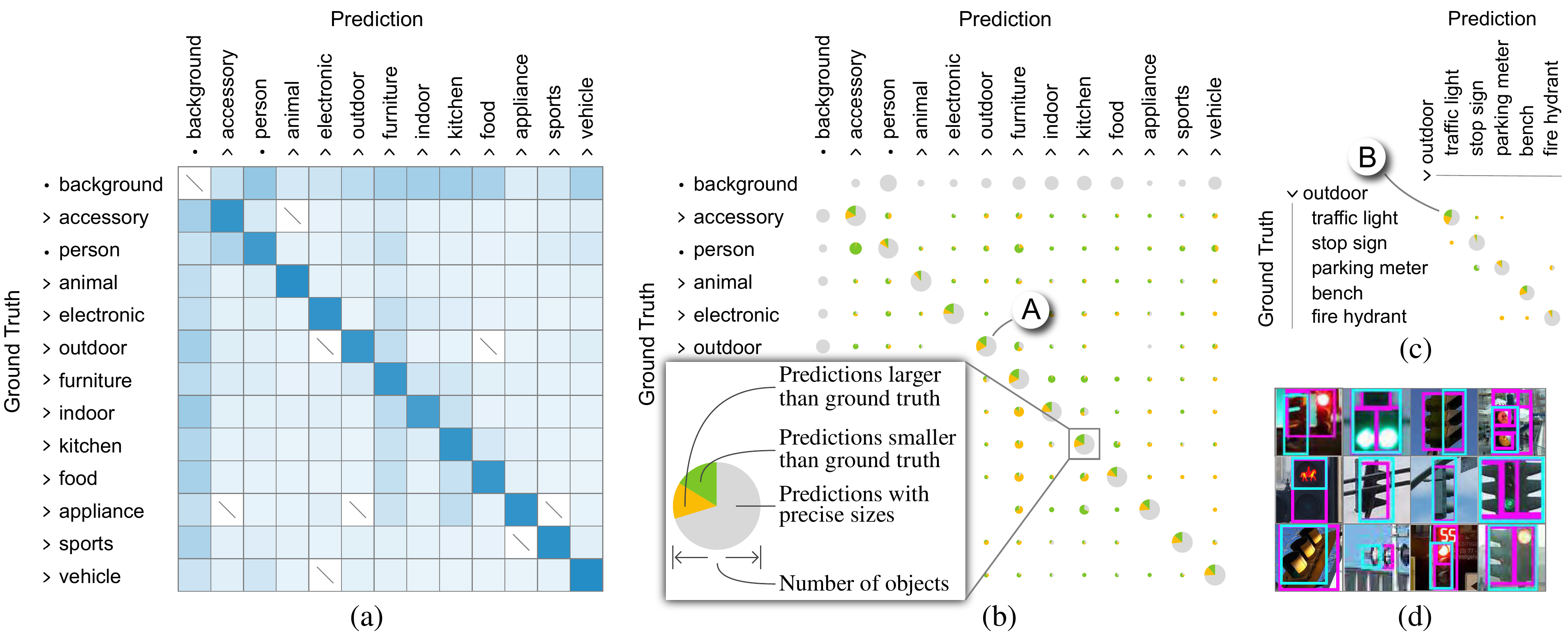}
    \vspace{-5mm}
    \caption{The matrix-based visualization of the COCO dataset in (a) the confusion mode and (b) the size mode; (c) the sub-matrix of super-class ``outdoor;'' (d) the predictions on the objects of ``traffic light.''}
    \label{fig:det-matrices}
\end{figure*}

\subsection{Table Visualization}
To help users identify problematic subsets, a frequent pattern mining-based search method is first developed to mine the candidate subsets.
Then a table visualization is employed to convey these subsets in terms of their attributes.
The users can rank these subsets by one attribute or the combination of multiple attributes to find problematic subsets.
\looseness=-1

\myparagraph{Frequent pattern mining-based search}.
Since it is computationally infeasible to evaluate model performance on all possible subsets, a search method to find candidate subsets is required. 
The state-of-the-art subset search method, DivExplorer~\cite{pastor2021looking,zhang2022sliceteller}, employs a 
frequent pattern mining-based search method~\cite{han2007frequent}
to find the subsets that meet a minimum subset size.
Then it ranks the subsets based on model performance on each of these subsets.
With this ranking, users can identify the subsets where the model performs poorly.
This method works well for classification tasks as it supports the search of the subsets sliced along discrete attributes.
To apply this method to detection and segmentation tasks, 
the continuous attributes need to be discretized.
Accordingly, we first employ the equal frequency discretization method~\cite{wong1987synthesizing} to divide the considered continuous attribute into $d$ intervals, each of which contains a similar number of samples/objects.
This discretization method is utilized because it well balances robustness and accuracy, and is one of the most efficient discretization methods~\cite{boulle2006modl}.
In our implementation, the minimum subset size is set as ${\beta}S_c$ to avoid the selection of small subsets, which contribute little to overall model performance.
Here, $S_c$ is the number of samples/objects in the class being explored. 
$\beta$ is set as 0.1 by default, and the user can adjust it according to the task at hand. 
$d$ is set as $1/\beta$ to ensure each interval contains approximately $S/d={\beta}S$ samples/objects,
where $S$ is the number of samples/objects in the dataset.\looseness=-1

\myparagraph{Identifying problematic subsets}.
We employ an interactive table visualization to visualize the candidate subsets and help identify the problematic ones.
In the table, each row represents a subset with all its attributes, such as the precision and average object size (Fig.~\ref{fig:teaser}(c)).
The discrete attributes are displayed as text, and the continuous attributes are displayed as bar items. 
We provide several interactions, such as filtering and ranking, to explore the subsets.
For example, users can select the subsets of a specific class by filtering, 
or rank the subsets by one attribute or the combination of multiple attributes.
Following Lineup~\cite{gratzl2013lineup}, the combination is achieved by dragging the header of a column onto the header of another column.

\subsection{Grid Visualization}
To enable efficient examination of the relevant samples, a grid visualization is employed because of its effectiveness in exploring image content~\cite{chen2020oodanalyzer, chen2022towards, rottmann2023mosaicsets}.
The cells of the grid display selected samples with their predictions.
To support real-time exploration, we utilize the $k$NN-based grid layout algorithm~\cite{chen2020oodanalyzer} to determine the position of each sample within the grid.
The algorithm preserves the proximity between samples
by first projecting them as a set of 2D points with t-SNE~\cite{maaten2008visualizing},
and then matching these points with the cells by solving a linear assignment problem.
To clearly show the detected/segmented objects,
we crop the images and present them in the corresponding cells.
\looseness=-1

\subsection{Interactive Model Evaluation}
The three coordinated visualizations work together to support an interactive model evaluation 
from a global overview to individual samples.
The matrix-based visualization provides users with an overview of model performance and helps identify the matrix cells with errors, such as classification errors.
The samples in such cells are then examined 
in the grid visualization to help users analyze the main causes of the errors.
If the causes for the errors are challenging to discern in the grid visualization,
users can turn to the table visualization to analyze the causes at the subset level. 
During the analysis,
users can rank the subsets based on different attributes, or utilize the Scented Widgets (\glyph[2]{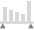})~\cite{willett2007scented} in Fig.~\ref{fig:teaser}(a) to select samples of interest.
With the identified causes of the errors, users can then make informed improvements to the model and/or the associated data.

\section{Case Studies}
To demonstrate the effectiveness of {\sys} for evaluating and improving different computer vision models (\textbf{T1}),
we conducted two case studies on object detection and instance segmentation.
Prior to the case studies, we briefly introduced {\sys} to the experts.
The visualizations of {\sys} were designed to be simple and familiar, allowing the experts to quickly understand its concepts and interactions within 20 minutes.
Throughout the case studies, we followed the pair analytics protocol~\cite{arias2011pair}, in which the experts led the exploration and analysis, and we navigated the tool.
This collaborative approach was chosen to enable the experts to fully focus on their analysis tasks, leading to enhanced efficiency and effectiveness.

\subsection{Object Detection on COCO dataset}
We collaborated with two experts ($E_1$ and $E_2$) to evaluate a state-of-the-art object detection model, DINO~\cite{zhang2022dino}.
$E_1$ is a Ph.D. student who developed DINO with his teammates.
$E_2$ is a researcher from a technology company who developed several object detection models and integrated them into their products.
A popular dataset for object detection, the COCO dataset~\cite{lin2014microsoft}, is utilized in this study.
This dataset consists of 118,287 training samples with 849,947 objects and 5,000 test samples with 36,335 objects.
The objects belong to 80 classes divided into 12 super-classes.
DINO with the ResNet-50 backbone~\cite{he2016deep} achieves mAPs of $65.2\%$ and $50.8\%$ on the training and test samples, respectively.
Although this model improved the mAP on COCO compared with previous object detection models,
the experts would like to examine what limited it from achieving a better performance.

Following common practices, the experts evaluated the model on the test samples.
They began the analysis by examining overall performance in the matrix-based visualization.
The matrix was initially in the confusion mode with row normalization, showing the confusion between 12 super-classes.
Most of the off-diagonal cells had very light colors (Fig.~\ref{fig:det-matrices}(a)), 
which indicated that the model had a high accuracy in classifying the objects of different super-classes.
Considering the high classification accuracy but low mAP, 
the experts suspected that the model classified objects well but failed to localize them accurately.
As localization concerns both the size errors and shifted directions of the predictions~\cite{girshick2015fast}, 
they decided to evaluate the model from these two perspectives.
$E_1$ focused on analyzing the sizes of predicted objects,
and $E_2$ focused on analyzing the shifted directions of predicted objects.

\subsubsection{Diagnosing Size Issues}
\myparagraph{Performance overview (T2)}.
To investigate the potential size errors in the predictions,
$E_1$ switched to the size mode.
In the matrix, he observed that all the large pie charts with large green and yellow sectors were on the diagonal.
This indicated that many objects were classified correctly but localized with size errors.
To investigate the cause of the size issue,
$E_1$ decided to dive deeper into these diagonal cells.
The diagonal cell with the most size errors, super-class ``outdoor'' (Fig.~\ref{fig:det-matrices}A), was taken as an example to illustrate the idea.
The other diagonal cells can be analyzed in a similar way.
\looseness=-1

\myparagraph{Analyzing size issues in the subset of ``outdoor'' (T2, T3, T4)}.
To figure out which classes within the super-class ``outdoor'' contributed to the size errors, he expanded this cell to a sub-matrix (Fig.~\ref{fig:det-matrices}(c)).
In the sub-matrix,
he found the diagonal cell of ``traffic light'' contributed the majority of the size errors (Fig.~\ref{fig:det-matrices}B).
However, in the grid visualization, the predictions on the associated objects did not show clear reasons for these size errors (Fig.~\ref{fig:det-matrices}(d)). 
$E_1$ then turned to the table visualization to analyze performance on different subsets in this cell.

\begin{figure}[!b]
\vspace{-5mm}
    \centering
    \includegraphics[width=\linewidth]{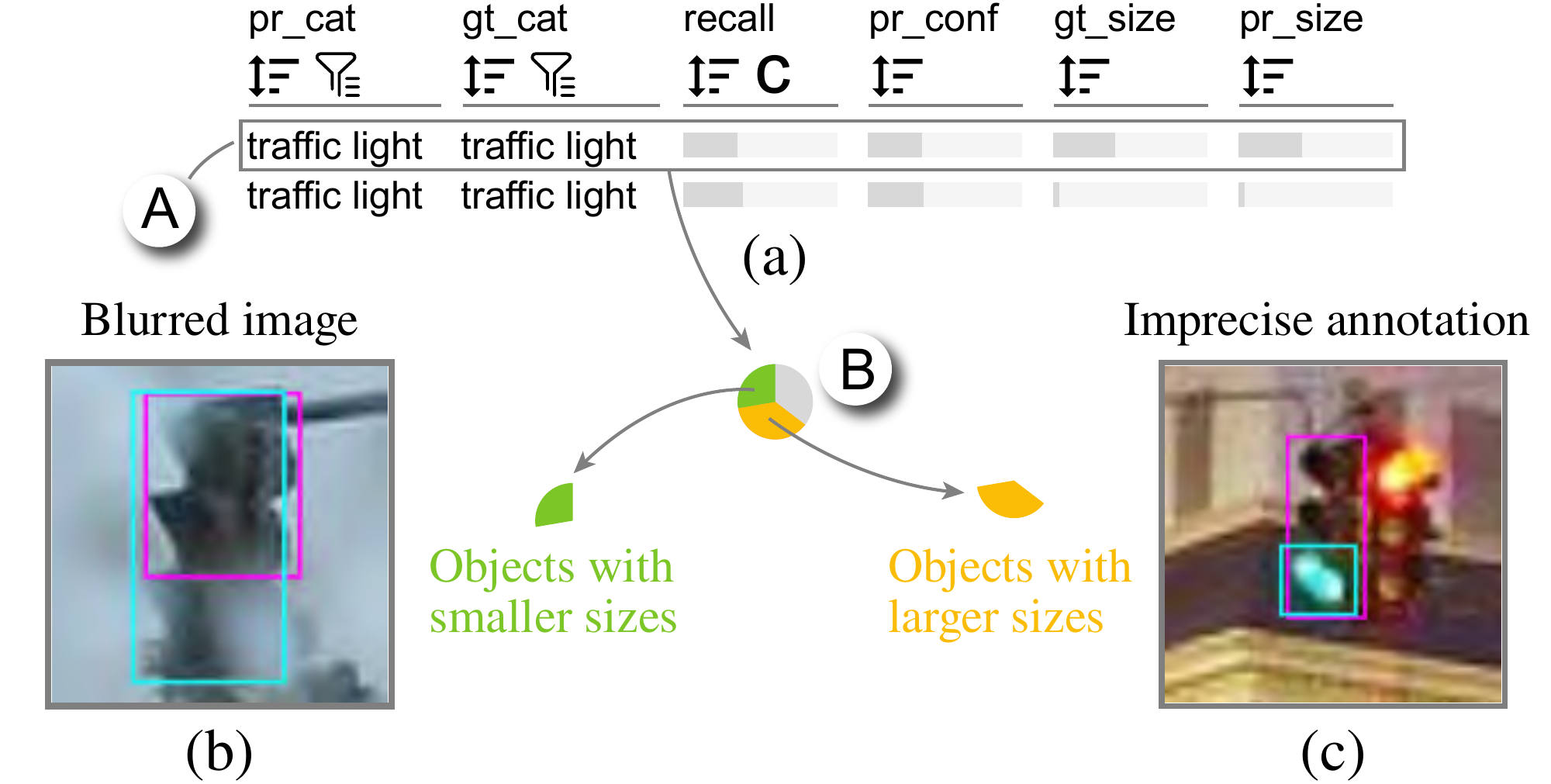}
    \vspace{-5mm}
    \caption{(a) The table visualization shows subsets of ``traffic light;'' (b) objects with smaller sizes; (c) objects with larger sizes.}
    \label{fig:case-traffic}
\end{figure}

To find problematic subsets in the table visualization,
$E_1$ sorted the subsets by the recall scores in ascending order. 
He identified many subsets with low recall scores (Fig.~\ref{fig:case-traffic}(a)).
$E_1$ selected the subset with the lowest recall score (Fig.~\ref{fig:case-traffic}A) and then turned to the matrix.
In the matrix, he clicked the green/yellow sector in the pie chart (Fig.~\ref{fig:case-traffic}B) to examine the predictions with smaller/larger sizes than ground truth in the grid visualization. 
\looseness=-1

From examining the predictions with smaller sizes, $E_1$ found that most of these predictions 
existed in blurred images (Fig.~\ref{fig:case-traffic}(b)).
$E_1$ examined the training samples and found only a few such blurred images.
He concluded that this was the main reason why the model did not perform well on these blurred images.
To tackle this issue,
$E_1$ applied Gaussian noise data augmentation to $4,139$ training samples with traffic lights.
After fine-tuning with the augmented samples,
the AP of ``traffic light'' was increased from $32.7\%$ to $33.1\%$.

From examining the predictions with larger sizes, $E_1$ found that 
the model had already made precise predictions on these objects (Fig.~\ref{fig:case-traffic}(c)).
However, some annotations were imprecise, which caused small overlaps between predictions and annotations.
$E_1$ continued to examine other subsets of ``traffic light'' and found similar issues caused by imprecise annotations.
He commented that the imprecise annotations in the test samples would mislead the model evaluation.
Interested in assessing actual performance on ``traffic light,''
$E_1$ hired annotators to re-annotate the bounding boxes of ``traffic light'' on test samples.
$E_1$ then re-evaluated the model using the re-annotated test samples. 
The AP of ``traffic light'' was increased from $33.1\%$ to $42.4\%$.

\begin{figure}[!t]
    \centering
    \includegraphics[width=\linewidth]{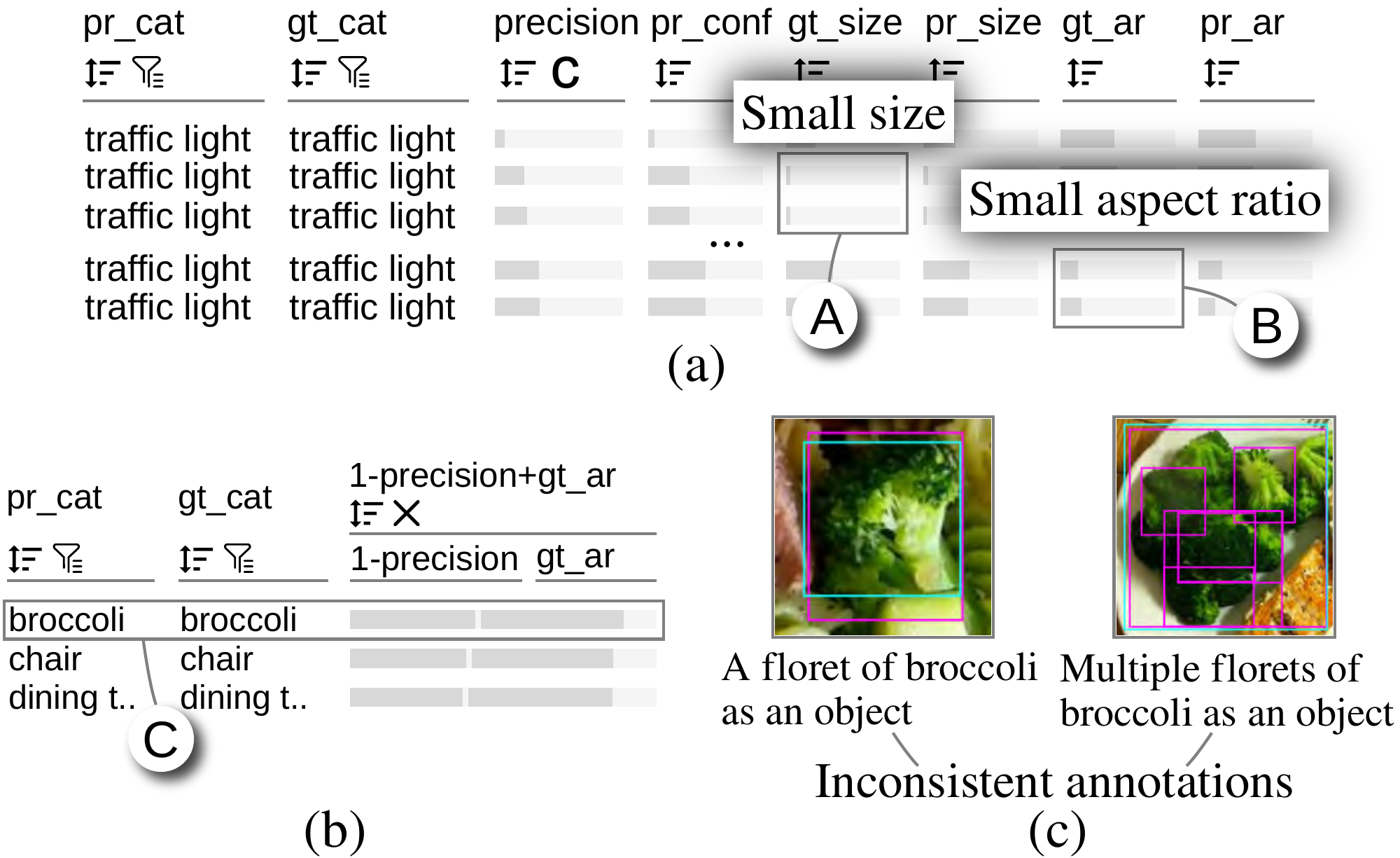}
    \vspace{-5mm}
    \caption{The subsets ranked by (a) the precision score and (b) the combination of the precision score and the ground-truth aspect ratio; (c) inconsistent annotations of ``broccoli.''}
    \label{fig:case-subsets}
\vspace{-5mm}
\end{figure}

After exploring the subsets with low recall scores,
$E_1$ was also interested in the subsets of ``traffic light'' with low precision scores.
He sorted the subsets by the precision scores in ascending order (Fig.~\ref{fig:case-subsets}(a))
and noticed that 
many subsets with low precision scores contained the objects of 
small sizes (Fig.~\ref{fig:case-subsets}A) or small aspect ratios (Fig.~\ref{fig:case-subsets}B).
Here, the aspect ratio of an object is 
the ratio between the minimum and maximum of the width and the height, 
which ranges between 0 and 1~\cite{hou2022shape}.
$E_1$ explained that this is reasonable because these characteristics represented some extremely hard objects in detection,
and therefore would lead to a low precision score in predictions.
Conversely, objects of large sizes or large aspect ratios were usually detected with high precision scores because they were easy to detect. 
Inspired by this finding,
$E_1$ would like to examine the subsets that were low in precision but without such hard objects.
Next, we took the subsets with low precision scores but large aspect ratios as examples for demonstration.

\myparagraph{Analyzing other low-precision subsets (T3, T4)}.
$E_1$ re-ranked the subsets by combining the precision score and the aspect ratio of ground truth to find some subsets with low precision scores but large aspect ratios (Fig.~\ref{fig:case-subsets}(b)).
Among the subsets, a subset of ``broccoli'' was ranked at the top (Fig.~\ref{fig:case-subsets}C).
He examined its associated objects and saw many ``broccoli'' objects were annotated with a confusing criterion (Fig.~\ref{fig:case-subsets}(c)).
Some contained only one floret of broccoli,
while others might include two or more florets of broccoli.
From these findings, 
he realized that the inconsistent annotations caused some correct predictions considered to be wrong during model evaluation.
Similar patterns were also found in other food-related classes, including ``apple,'' ``orange,'' ``banana,'' and ``carrot.''
For more accurate model evaluation,
$E_1$ decided to re-annotate these five classes with a consistent criterion.
After the re-annotation, the APs of ``broccoli,'' ``apple,'' ``banana,'' ``orange,'' and ``carrot'' were increased from 
$26.9\%$ to $62.1\%$, $30.3\%$ to $38.7\%$,
$32.9\%$ to $51.0\%$, $38.7\%$ to $78.4\%$,
$29.3\%$ to $44.4\%$, respectively.
\looseness=-1

\subsubsection{Diagnosing Direction Issues}
\begin{figure}[!t]
    \centering
    \includegraphics[width=\linewidth]{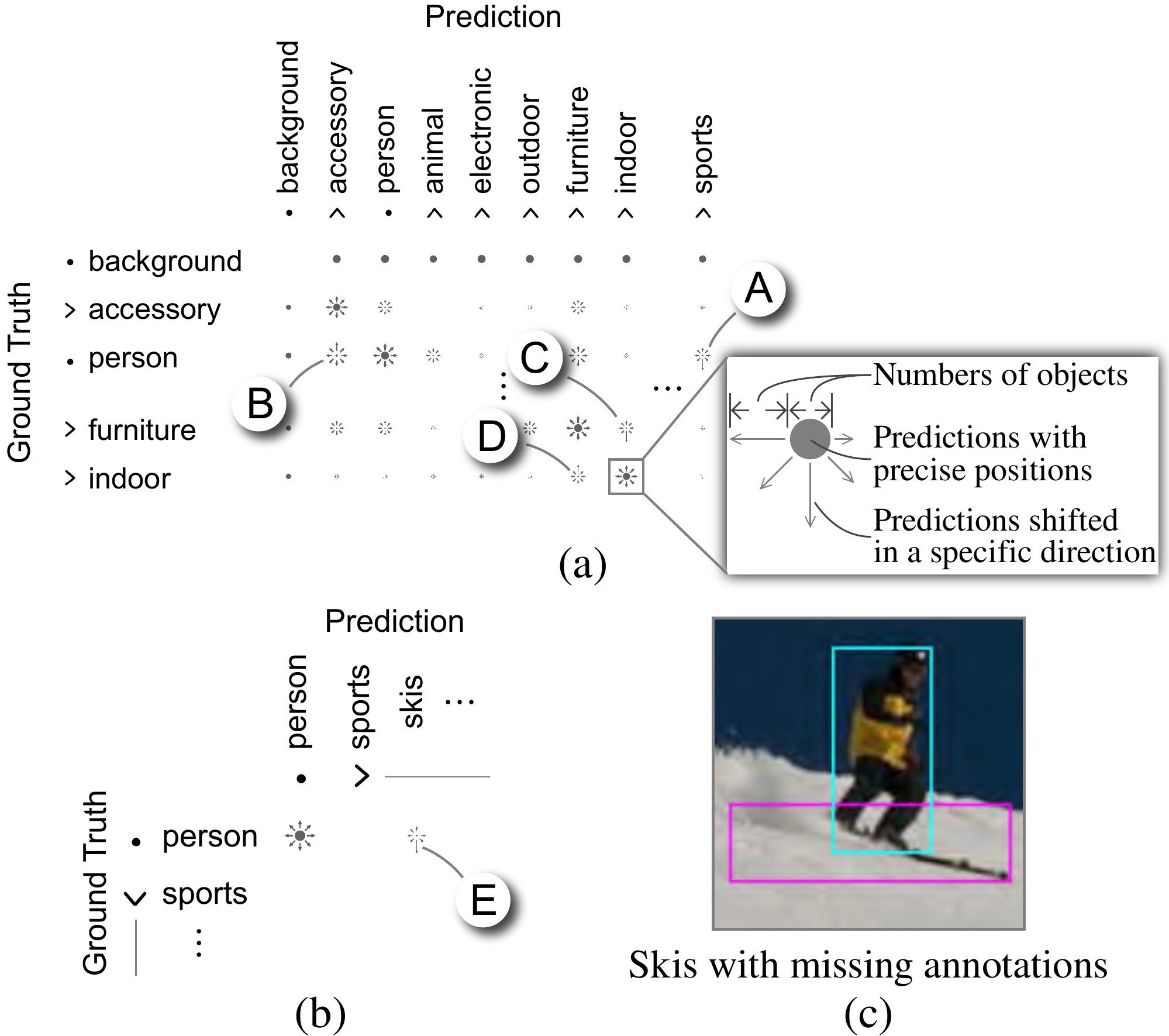}
    \vspace{-5mm}
    \caption{(a) The matrix-based visualization in the direction mode; (b) the sub-matrix of ``person'' and ``sports;'' (c) missing annotations in ``skis.''}
    \label{fig:case-direction}
    \vspace{-5mm}
\end{figure}

\myparagraph{Performance overview (T2)}.
$E_2$ continued the evaluation of localization performance in terms of shifted directions.
He examined the matrix in the direction mode.
In the matrix (Fig.~\ref{fig:case-direction}(a)),
he found four cells with apparently longer arrows in specific directions (Figs.~\ref{fig:case-direction}A,~\ref{fig:case-direction}B,~\ref{fig:case-direction}C, and~\ref{fig:case-direction}D).
He decided to examine them one by one.

\myparagraph{Analyzing the imprecise direction issue in class ``person'' (T2, T4)}.
$E_2$ began his analysis on cell A (Fig.~\ref{fig:case-direction}A),
where many predicted objects were shifted downward.
To determine the reason for the direction errors,
he expanded the cell to a sub-matrix (Fig.~\ref{fig:case-direction}(b)).
In the sub-matrix,
he noticed cell E (Fig.~\ref{fig:case-direction}E), where ``person'' was confused with ``skis,'' contributed the majority of the direction errors.
A large proportion of predictions in this cell were shifted downward in position.
He clicked the downward arrow to further check the associated objects.
In the grid visualization,
he noticed that the predicted ``skis'' were precisely localized,
but the corresponding annotations were absent (Fig.~\ref{fig:case-direction}(c)).
As a result, the predictions were matched with the ground truth ``person.''
$E_2$ decided to add the missing annotations of ``skis'' in test samples.
After that, the AP of ``skis'' was increased from $35.4\%$ to $36.8\%$.

Similarly, $E_2$ examined the other cells with longer arrows in specific directions (Figs.~\ref{fig:case-direction}B,~\ref{fig:case-direction}C and \ref{fig:case-direction}D).
Analyzing the predictions,
he discovered similar issues as ``skis''
and decided to add the missing annotations of ``potted plant'' and ``vase'' in test samples.
After that, the APs of ``potted plant'' and ``vase'' were increased from $33.6\%$ to $37.0\%$, and $44.1\%$ to $46.2\%$, respectively.

In summary, $E_1$ and $E_2$ applied Gaussian noise data augmentation to $4,139$ training samples with traffic lights and re-annotated the objects of nine classes.
The overall mAP on the re-annotated test samples was improved from $50.8\%$ to $52.5\%$.
$E_2$ was impressed by the improvement and thus conducted a post-analysis on the re-annotated test samples.

\subsubsection{Post Analysis}

\myparagraph{More accurately evaluating model performance}.
The previous study has shown that larger models usually achieve higher mAPs than small models on the original COCO test samples~\cite{fang2022eva}.
$E_2$ sought to determine whether they truly perform better or merely overfit the annotation errors.
Therefore, $E_2$ evaluated performance of InternImage~\cite{wang2022internimage}, a large pre-trained model that achieves the highest mAP ($65.0\%$) on the original test samples.
On the re-annotated test samples, the gain in mAP was $2.3\%$, which exceeded that of DINO ($1.8\%$).
This showed that the higher performance on large models was not due to overfitting annotation errors even though the numbers of their parameters were very large.
This is also verified by the recent research~\cite{yang2022focal}.

Meanwhile, $E_2$ noted that the current annotations might not be accurate enough to evaluate object detection models.
``Considering the $2.3\%$ mAP improvement by revising the annotations of only nine classes, I expect the gain to reach around $15.0\%$ if the other 71 classes were re-annotated.
It makes me largely underestimate performance of object detection models.'' $E_2$ said. 
To accurately evaluate models, $E_2$ emphasized the importance of re-annotating the remaining 71 classes. 
The findings in the case study provide guidance for the re-annotation process.
For example, when numerous objects of the same class are present in close proximity and delineating them is difficult even for humans (\eg, the broccoli in Fig.~\ref{fig:case-subsets}(c)), 
these objects can be annotated as a whole with the label ``is crowd'' to avoid inaccurate annotations.

\myparagraph{Distilling large models}.
As inaccurate annotations were identified in the test samples, 
$E_2$ suspected the annotations of the training samples were also inaccurate.
The inaccurate annotations of the training images will degrade model performance, especially for small machine learning models running on resource-limited devices~\cite{nakkiran2021deep}.
As re-annotating the training samples is prohibitively expensive, $E_2$ suggested using distillation to enhance the performance of small models.
Specifically, a large model is trained on the training samples and then utilized to detect the objects in all the training samples.
Then these detected objects are used to train the small models.
To illustrate the idea, $E_2$ used DINO with Swin backbone~\cite{liu2021swin} as the large model and DINO with ResNet-50 backbone~\cite{he2016deep} as the small model for distillation.
On the re-annotated test samples, the mAP was improved from $52.5\%$ to $53.4\%$.
``The improvement is remarkable, particularly because it required no additional annotation efforts.'' $E_2$ commented.
\looseness=-1

\subsection{Instance Segmentation on iSAID dataset}
In this case study, we invited the expert $E_3$ to evaluate an instance segmentation model on the iSAID dataset~\cite{waqas2019isaid}, an aerial image dataset for instance segmentation.
This dataset contains 36,038 training samples with 716,640 objects and 11,752 test samples with 233,625 objects.
The objects belong to 15 classes divided into two super-classes, ``transport'' and ``land.''
$E_3$ employed CATNet~\cite{liu2021catnet}, a state-of-the-art instance segmentation model for aerial images.
With the ResNet-50 backbone~\cite{he2016deep}, it achieved an mAP of $51.7\%$ on the training samples and an mAP of $39.1\%$ on the test samples. 
$E_3$ wanted to improve model performance, so he used {\sys} to evaluate the model on the test samples.
Here, we take the super-class ``transport'' as an example to illustrate the idea. 
\looseness=-1

\myparagraph{Performance overview (T2)}.
Initially, the matrix was in the confusion mode with the super-class ``transport'' expanded,
displaying the confusion between seven classes (Fig.~\ref{fig:teaser}(b)).
In the matrix, he discovered several regions of interest: 
1) in region A, ``large vehicle'' and ``small vehicle'' were confused with each other (Fig.~\ref{fig:teaser}A);
2) in region B, many objects of ``ship,'' ``large vehicle,'' ``small vehicle,'' ``storage tank,'' and ``helicopter'' failed to be segmented (Fig.~\ref{fig:teaser}B);
3) in region C, many helicopters were misclassified as planes (Fig.~\ref{fig:teaser}C); and
4) in region D,  some backgrounds were misclassified as ``large vehicle'' and ``small vehicle'' (Fig.~\ref{fig:teaser}D).
Since the classification and the segmentation were interrelated in this multi-task scenario,
$E_3$ further examined the sizes and the shifted directions of the segmented objects with the size mode and direction mode, respectively.
In the size mode, he found two regions of interest:
in region E, some objects of ``small vehicle'' were confused with ``large vehicle'' (Fig.~\ref{fig:teaser}E); and in region F, some objects of ``harbor'' were correctly classified but with size errors (Fig.~\ref{fig:teaser}F).
However, $E_3$ found no obvious patterns in the direction mode.
He explained, ``This is reasonable because aerial images are insensitive to direction changes.''
$E_3$ decided to analyze each identified region of interest separately.

\myparagraph{Analyzing misclassification in ``large vehicle'' and ``small vehicle'' (T2, T4)}.
$E_3$ first analyzed region A, where ``large vehicle'' and ``small vehicle'' were confused with each other (Fig.~\ref{fig:teaser}A).
To understand what caused the confusion, he switched to the size mode and observed that in 
cell E (Fig.~\ref{fig:teaser}E),
the yellow sector of the pie chart occupied a large proportion.
This indicated that many segmented small vehicles were larger than ground truth.
To investigate this further, $E_3$ clicked the yellow sector to check the associated objects (Fig.~\ref{fig:teaser}(d)).
He noticed that the small vehicles were so small that most of the predictions masked two or more small vehicles as a whole.
Such wrong segmentation results were also found in the training samples.
$E_3$ hypothesized that the model was poor at predicting such small objects.
To verify this, he selected the small objects by filtering and found that most of these objects belonged to five classes: ``ship,'' ``storage tank,'' ``small vehicle,'' ``large vehicle,'' and ``helicopter.''
Many objects of these five classes were misclassified (Fig.~\ref{fig:teaser}A) or failed to be segmented (Fig.~\ref{fig:teaser}B).
Upon this observation, he examined the model and found that the images were downsampled in the model.
The small objects were hard to be distinguished in the downsampled images, and thus led to the wrong segmentation.
To address this issue, $E_3$ increased the resolution of the images from $512\times512$ to $1024\times1024$ and retrained the model.
Subsequently, the APs of ``ship,'' ``storage tank,'' ``small vehicle,'' ``large vehicle,'' and ``helicopter'' were increased from
$49.9\%$ to $51.5\%$, $39.9\%$ to $42.8\%$,
$16.5\%$ to $19.2\%$, $39.5\%$ to $41.6\%$,
$6.3\%$ to $7.5\%$, respectively,
and the overall mAP was increased from $39.1\%$ to $40.3\%$.
\looseness=-1

\begin{figure}[t]
    \centering
    \includegraphics[width=\linewidth]{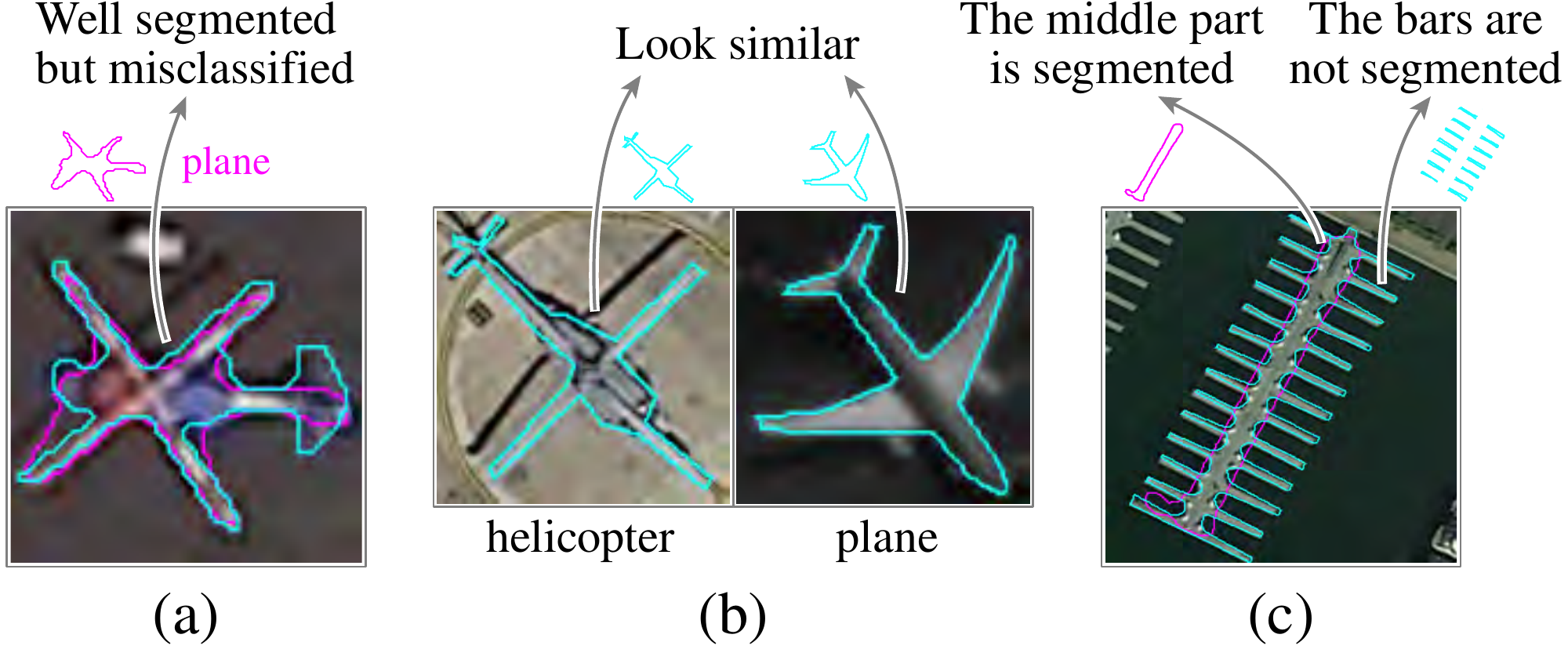}
    \vspace{-6mm}
    \caption{(a) A helicopter is well segmented but misclassified; (b) a helicopter and a plane share a similar appearance; (c) the middle part of a harbor is segmented, but its bars are not.}
    \label{fig:case-seg}
\vspace{-5mm}
\end{figure}

\myparagraph{Analyzing misclassification in ``helicopter'' and background (T2, T4)}.
$E_3$ proceeded to analyze region C and found that some objects of ``helicopter'' was misclassified as ``plane.''
Examining the associated objects in the grid visualization, he observed that these helicopters were well segmented but still misclassified (Fig.~\ref{fig:case-seg}(a)).
After examining helicopters and planes in the training samples, $E_3$ commented, ``This misclassification occurs because helicopters and planes have a similar appearance in aerial images (Fig.~\ref{fig:case-seg}(b)), and the training samples contain more instances of planes (19,720) than helicopters (1,402).''
To address this issue, $E_3$ first collected 1,077 aerial images with 1,561 helicopters,
then applied the widely-used Copy-Paste augmentation strategy~\cite{ghiasi2021simple} to generate more helicopters in the training samples.
The augmentation was carried out by pasting the helicopters from their original images into other images, which resulted in a total of 5,063 helicopters.
\looseness=-1

Next, $E_3$ analyzed region D, where some backgrounds were misclassified as ``large vehicle'' or ``small vehicle'' (Fig.~\ref{fig:teaser}D).
$E_3$ considered this was reasonable since vehicles were small and obscure in aerial images, making them difficult to be distinguished
from the background by the model.
This was also consistent with the previous studies~\cite{waqas2019isaid, cao2020d2det}.
As this misclassification was a long-standing problem in instance segmentation, $E_3$ decided to explore it further in future research.

After fine-tuning the model with the augmented samples of ``helicopter'', the AP of ``helicopter'' was increased from $7.5\%$ to $10.4\%$.
The overall mAP was increased from $40.3\%$ to $40.4\%$.

\myparagraph{Analyzing the imprecise size issue in ``harbor'' (T2, T4)}.
$E_3$ continued to analyze the diagonal cell of ``harbor'' (Fig.~\ref{fig:teaser}F).
The green sector of this cell indicated that many harbors were predicted with a smaller size.
Intrigued by this observation, 
he clicked the green sector to check the associated objects in the grid visualization.
He found that the middle parts of the harbors were correctly segmented, but the bars around them were not (Fig.~\ref{fig:case-seg}(c)).
Similar patterns were also found in the training samples.
From the finding, 
he realized that the currently used binary cross entropy loss function is area-based, and the model with this loss function was unable to precisely segment the bars due to their small sizes.
It was necessary to consider the boundaries of harbors during training. 
Additionally, $E_3$  found two other classes, ``plane'' and ``helicopter,'' with the similar issue of complex boundaries. 
To address this issue, he combined a boundary-based loss~\cite{kervadec2021boundary} with the original binary cross entropy loss for the three classes.
After fine-tuning the model, the APs of ``harbor,'' ``plane,'' and ``helicopter'' were increased from $29.6\%$ to $31.7\%$, $51.5\%$ to $52.6\%$, and $10.4\%$ to $11.3\%$, respectively.
This resulted in an overall mAP increase to $40.6\%$.

In summary, $E_3$:
1) increased the resolution of all images; 
2) added $1,077$ samples with $1,561$ helicopters and utilized Copy-Paste augmentation to obtain a total of $5,063$ helicopters; 
3) and added a boundary-based loss for ``harbor,'' ``plane,'' and ``helicopter.''
The overall mAP was improved from $39.1\%$ to $40.6\%$.
\looseness=-1

\section{Expert Feedback and Discussion}
\label{sec:discuss}
Following the case studies, we conducted six interviews to gather feedback from a group of experts, which included the three experts ($E_1$, $E_2$, $E_3$) who collaborated with us in the case studies and three additional experts we invited ($E_4$, $E_5$, $E_6$).
The newly invited experts were Ph.D.~students who had over two years of experience in computer vision research.
Each interview lasted between 40 to 65 minutes.
Overall, the expert feedback was positive regarding the usability of {\sys}. 
However, some limitations were also identified, highlighting areas requiring further investigation in the future.

\subsection{Usability}

\noindent\textbf{Generalization to other tasks}.
In the current implementation, {\sys} supports three main tasks in computer vision: classification, object detection, and instance segmentation.
According to the experts, {\sys} can also be utilized for evaluating models in other computer vision tasks.
Here we take the visual generation task as an example to illustrate the extension idea.
This task involves encoding training samples into latent vectors and then reconstructing the original training samples.
When evaluating models in such tasks, the experts are interested in analyzing the reconstruction errors of samples.
To do the analysis, they can cluster samples and then analyze the errors between different clusters in the matrix-based visualization.
Meanwhile, the table visualization and the grid visualization are directly applicable to exploring the subsets and samples of interest.
The semantic segmentation task can also be analyzed in {\sys} as a pixel-level classification task. 
However, since this task usually has a significant class imbalance, such a pixel-level analysis may hinder the poor performance analysis on minority classes. 
In addition, focusing only on pixel-level information results in a loss of object-level information (the shapes of objects), which is also critical for evaluating semantic segmentation models~\cite{he2021can}. Therefore, it is essential to study metrics that are insensitive to class imbalance (\eg, mIoU) and integrate pixel- and object-level information in {\sys}, which is a promising research topic in the future.
\looseness=-1

\noindent\textbf{Diagnosing causes of poor performance}.
All the experts commented positively on the design of the three modes in the matrix-based visualization, 
which enables the identification of the specific reasons for poor performance.
In the instance segmentation case study, using the three modes, $E_3$ successfully identified that the confusion between ``small vehicle'' and ``large vehicle'' was due to the small sizes of the ``small vehicle.''
Furthermore, $E_1$ noted that the table visualization facilitated diagnosing the different causes of poor performance in different subsets.
``Using the table visualization, I quickly identified that the size errors of class `traffic light' were caused by both the blurred images and the annotation errors in the test samples,'' he commented.\looseness=-1

\subsection{Limitations and Future Work}

\noindent\textbf{Identifying data subsets by semantic attributes}.
{\sys} currently slices the subsets along the low-level attributes of objects, such as sizes and aspect ratios.
It does not support identifying problematic subsets based on semantic attributes that describe visual appearances. 
For example, $E_2$ observed that their object detection models often failed to detect objects with some specific appearances, 
such as a person on a bridge.
However, such semantic attributes usually require expensive manual annotations~\cite{zhang2022sliceteller}.
One potential method is to use disentangled representation learning to extract candidate semantic attributes and then allow users to select meaningful ones~\cite{gou2020vatld, he2021can, wang2023drava}.
\looseness=-1

\noindent\textbf{Effective model comparison}.
Uni-Evaluator is currently designed for evaluating a single model. However, the experts also want to compare different models for various purposes~\cite{derose2020attention,sevastjanova2022visual}. 
For example, $E_4$ indicated that Transformer-based object detection models usually outperformed CNN-based object detection models,
whereas the latter often exhibited higher inference speeds in real-world applications~\cite{liu2022convnet}.
To exploit the strengths of both, $E_4$ wanted to compare their performance, identify what made the Transformer-based models perform better, and then integrate it into the CNN-based models.
Therefore, investigating how {\sys} can effectively support model comparison is a promising future research direction.

\section{Conclusion}

We present {\sys}, a visual analysis tool that supports a unified interactive model evaluation for computer vision tasks.
From a survey conducted with 151 computer vision experts, we distill three design goals for a unified model evaluation and then derive four tasks from the design goals.
Based on the derived tasks,
we propose a unified probability distribution method that models both continuous and discrete predictions in a unified manner.
With the unified probabilistic modeling,
we develop three coordinated visualizations to facilitate a comprehensive model evaluation from a global overview to detailed samples.
Two case studies are conducted to demonstrate the effectiveness of {\sys} in
improving the model and the associated data in object detection and instance segmentation.

\acknowledgments{
This work was supported by the National Natural Science Foundation of China under grants U21A20469, 61936002, and 92248303, the National Key R\&D Program of China under Grant 2020YFB2104100, grants from the Institute Guo Qiang, THUIBCS, and BLBCI, and in part by Tsinghua-Kuaishou Institute of Future Media Data.
The authors would like to thank Dr.~Xizhou Zhu, Dr.~Mengchen Liu, Guo-Ye Yang, Dr.~Liuyu Xiang, and Dr.~Xiaohan Wang for their contributions to the case studies, Lanxi Xiao for her valuable comments on the visualization design, Zhen Li for implementing parts of the matrix-based visualization and the filtering function, and Yiwei Hou for voicing our video. 
}

\bibliographystyle{abbrv-doi-hyperref}

\bibliography{reference}

\begin{thebibliography}{10}

\bibitem{abadi2016tensorflow}
M.~Abadi, A.~Agarwal, P.~Barham, E.~Brevdo, Z.~Chen, C.~Citro, G.~S. Corrado,
  A.~Davis, J.~Dean, M.~Devin, et~al.
\newblock Tensorflow: Large-scale machine learning on heterogeneous distributed
  systems.
\newblock {\em CoRR}, abs/1603.04467, 2016.
  \href{https://doi.org/10.48550/arXiv.1603.04467}
{doi: {{%
10\hspace{.1pt}\discretionary{.}{%
}{.}\hspace{.4pt}48550\discretionary{/}{%
}{/}arXiv\hspace{.1pt}\discretionary{.}{%
}{.}\hspace{.4pt}1603\hspace{.1pt}\discretionary{.}{%
}{.}\hspace{.4pt}04467}}}


\bibitem{alsallakh2014visual}
B.~Alsallakh, A.~Hanbury, H.~Hauser, S.~Miksch, and A.~Rauber.
\newblock Visual methods for analyzing probabilistic classification data.
\newblock {\em IEEE Transactions on Visualization and Computer Graphics},
  20(12):1703--1712, 2014. \href{https://doi.org/10.1109/TVCG.2014.2346660}
{doi: {{%
10\hspace{.1pt}\discretionary{.}{%
}{.}\hspace{.4pt}1109\discretionary{/}{%
}{/}TVCG\hspace{.1pt}\discretionary{.}{%
}{.}\hspace{.4pt}2014\hspace{.1pt}\discretionary{.}{%
}{.}\hspace{.4pt}2346660}}}


\bibitem{amershi2015modeltracker}
S.~Amershi, M.~Chickering, S.~M. Drucker, B.~Lee, P.~Simard, and J.~Suh.
\newblock {ModelTracker}: Redesigning performance analysis tools for machine
  learning.
\newblock In {\em Proceedings of the ACM CHI Conference on Human Factors in
  Computing Systems}, pp. 337--346. Seoul, 2015.
  \href{https://doi.org/10.1145/2702123.2702509}
{doi: {{%
10\hspace{.1pt}\discretionary{.}{%
}{.}\hspace{.4pt}1145\discretionary{/}{%
}{/}2702123\hspace{.1pt}\discretionary{.}{%
}{.}\hspace{.4pt}2702509}}}


\bibitem{arias2011pair}
R.~Arias-Hernandez, L.~T. Kaastra, T.~M. Green, and B.~Fisher.
\newblock Pair analytics: Capturing reasoning processes in collaborative visual
  analytics.
\newblock In {\em IEEE Hawaii International Conference on System Sciences}, pp.
  1--10. Koloa, Kauai, 2011. \href{https://doi.org/10.1109/HICSS.2011.339}
{doi: {{%
10\hspace{.1pt}\discretionary{.}{%
}{.}\hspace{.4pt}1109\discretionary{/}{%
}{/}HICSS\hspace{.1pt}\discretionary{.}{%
}{.}\hspace{.4pt}2011\hspace{.1pt}\discretionary{.}{%
}{.}\hspace{.4pt}339}}}


\bibitem{bar2001fast}
Z.~Bar-Joseph, D.~K. Gifford, and T.~S. Jaakkola.
\newblock Fast optimal leaf ordering for hierarchical clustering.
\newblock {\em Bioinformatics}, 17(suppl\_1):S22--S29, 2001.
  \href{https://doi.org/10.1093/bioinformatics/17.suppl_1.S22}
{doi: {{%
10\hspace{.1pt}\discretionary{.}{%
}{.}\hspace{.4pt}1093\discretionary{/}{%
}{/}bioinformatics\discretionary{/}{%
}{/}17\hspace{.1pt}\discretionary{.}{%
}{.}\hspace{.4pt}suppl\_1\hspace{.1pt}\discretionary{.}{%
}{.}\hspace{.4pt}S22}}}


\bibitem{bertucci2022dendromap}
D.~Bertucci, M.~M. Hamid, Y.~Anand, A.~Ruangrotsakun, D.~Tabatabai, M.~Perez,
  and M.~Kahng.
\newblock {DendroMap}: Visual exploration of large-scale image datasets for
  machine learning with treemaps.
\newblock {\em IEEE Transactions on Visualization and Computer Graphics},
  29(1):320--330, 2023. \href{https://doi.org/10.1109/TVCG.2022.3209425}
{doi: {{%
10\hspace{.1pt}\discretionary{.}{%
}{.}\hspace{.4pt}1109\discretionary{/}{%
}{/}TVCG\hspace{.1pt}\discretionary{.}{%
}{.}\hspace{.4pt}2022\hspace{.1pt}\discretionary{.}{%
}{.}\hspace{.4pt}3209425}}}


\bibitem{biewald2020experiment}
L.~Biewald.
\newblock {Weights \& Biases}.
\newblock \url{https://wandb.ai/site/}, 2020.

\bibitem{bilal2017convolutional}
A.~Bilal, A.~Jourabloo, M.~Ye, X.~Liu, and L.~Ren.
\newblock Do convolutional neural networks learn class hierarchy?
\newblock {\em IEEE Transactions on Visualization and Computer Graphics},
  24(1):152--162, 2018. \href{https://doi.org/10.1109/TVCG.2017.2744683}
{doi: {{%
10\hspace{.1pt}\discretionary{.}{%
}{.}\hspace{.4pt}1109\discretionary{/}{%
}{/}TVCG\hspace{.1pt}\discretionary{.}{%
}{.}\hspace{.4pt}2017\hspace{.1pt}\discretionary{.}{%
}{.}\hspace{.4pt}2744683}}}


\bibitem{bolya2020tide}
D.~Bolya, S.~Foley, J.~Hays, and J.~Hoffman.
\newblock {TIDE}: A general toolbox for identifying object detection errors.
\newblock In {\em Proceedings of the European Conference on Computer Vision},
  pp. 558--573. Glasgow, 2020.
  \href{https://doi.org/10.1007/978-3-030-58580-8_33}
{doi: {{%
10\hspace{.1pt}\discretionary{.}{%
}{.}\hspace{.4pt}1007\discretionary{/}{%
}{/}978\discretionary{%
}{-}{-}3\discretionary{%
}{-}{-}030\discretionary{%
}{-}{-}58580\discretionary{%
}{-}{-}8\_33}}}


\bibitem{borji2019empirical}
A.~Borji and S.~M. Iranmanesh.
\newblock Empirical upper bound in object detection and more.
\newblock {\em CoRR}, abs/1911.12451, 2019.
  \href{https://doi.org/10.48550/arXiv.1911.12451}
{doi: {{%
10\hspace{.1pt}\discretionary{.}{%
}{.}\hspace{.4pt}48550\discretionary{/}{%
}{/}arXiv\hspace{.1pt}\discretionary{.}{%
}{.}\hspace{.4pt}1911\hspace{.1pt}\discretionary{.}{%
}{.}\hspace{.4pt}12451}}}


\bibitem{boulle2006modl}
M.~Boull{\'e}.
\newblock {MODL}: {A Bayes} optimal discretization method for continuous
  attributes.
\newblock {\em Machine Learning}, 65:131--165, 2006.
  \href{https://doi.org/10.1007/s10994-006-8364-x}
{doi: {{%
10\hspace{.1pt}\discretionary{.}{%
}{.}\hspace{.4pt}1007\discretionary{/}{%
}{/}s10994\discretionary{%
}{-}{-}006\discretionary{%
}{-}{-}8364\discretionary{%
}{-}{-}x}}}


\bibitem{cao2020d2det}
J.~Cao, H.~Cholakkal, R.~M. Anwer, F.~S. Khan, Y.~Pang, and L.~Shao.
\newblock {D2Det}: Towards high quality object detection and instance
  segmentation.
\newblock In {\em Proceedings of the IEEE/CVF Conference on Computer Vision and
  Pattern Recognition}, pp. 11485--11494. Seattle, 2020.
  \href{https://doi.org/10.1109/cvpr42600.2020.01150}
{doi: {{%
10\hspace{.1pt}\discretionary{.}{%
}{.}\hspace{.4pt}1109\discretionary{/}{%
}{/}cvpr42600\hspace{.1pt}\discretionary{.}{%
}{.}\hspace{.4pt}2020\hspace{.1pt}\discretionary{.}{%
}{.}\hspace{.4pt}01150}}}


\bibitem{chen2021interactive}
C.~Chen, Z.~Wang, J.~Wu, X.~Wang, L.-Z. Guo, Y.-F. Li, and S.~Liu.
\newblock Interactive graph construction for graph-based semi-supervised
  learning.
\newblock {\em IEEE Transactions on Visualization and Computer Graphics},
  27(9):3701--3716, 2021. \href{https://doi.org/10.1109/TVCG.2021.3084694}
{doi: {{%
10\hspace{.1pt}\discretionary{.}{%
}{.}\hspace{.4pt}1109\discretionary{/}{%
}{/}TVCG\hspace{.1pt}\discretionary{.}{%
}{.}\hspace{.4pt}2021\hspace{.1pt}\discretionary{.}{%
}{.}\hspace{.4pt}3084694}}}


\bibitem{chen2022towards}
C.~Chen, J.~Wu, X.~Wang, S.~Xiang, S.-H. Zhang, Q.~Tang, and S.~Liu.
\newblock Towards better caption supervision for object detection.
\newblock {\em IEEE Transactions on Visualization and Computer Graphics},
  28(4):1941--1954, 2022. \href{https://doi.org/10.1109/TVCG.2021.3138933}
{doi: {{%
10\hspace{.1pt}\discretionary{.}{%
}{.}\hspace{.4pt}1109\discretionary{/}{%
}{/}TVCG\hspace{.1pt}\discretionary{.}{%
}{.}\hspace{.4pt}2021\hspace{.1pt}\discretionary{.}{%
}{.}\hspace{.4pt}3138933}}}


\bibitem{chen2020oodanalyzer}
C.~Chen, J.~Yuan, Y.~Lu, Y.~Liu, H.~Su, S.~Yuan, and S.~Liu.
\newblock {OoDAnalyzer}: Interactive analysis of out-of-distribution samples.
\newblock {\em IEEE Transactions on Visualization and Computer Graphics},
  27(7):3335--3349, 2021. \href{https://doi.org/10.1109/TVCG.2020.2973258}
{doi: {{%
10\hspace{.1pt}\discretionary{.}{%
}{.}\hspace{.4pt}1109\discretionary{/}{%
}{/}TVCG\hspace{.1pt}\discretionary{.}{%
}{.}\hspace{.4pt}2020\hspace{.1pt}\discretionary{.}{%
}{.}\hspace{.4pt}2973258}}}


\bibitem{derose2020attention}
J.~F. DeRose, J.~Wang, and M.~Berger.
\newblock Attention flows: Analyzing and comparing attention mechanisms in
  language models.
\newblock {\em IEEE Transactions on Visualization and Computer Graphics},
  27(2):1160--1170, 2021. \href{https://doi.org/10.1109/TVCG.2020.3028976}
{doi: {{%
10\hspace{.1pt}\discretionary{.}{%
}{.}\hspace{.4pt}1109\discretionary{/}{%
}{/}TVCG\hspace{.1pt}\discretionary{.}{%
}{.}\hspace{.4pt}2020\hspace{.1pt}\discretionary{.}{%
}{.}\hspace{.4pt}3028976}}}


\bibitem{fang2022eva}
Y.~Fang, W.~Wang, B.~Xie, Q.~Sun, L.~Wu, X.~Wang, T.~Huang, X.~Wang, and
  Y.~Cao.
\newblock {EVA}: Exploring the limits of masked visual representation learning
  at scale.
\newblock In {\em Proceedings of the IEEE/CVF Conference on Computer Vision and
  Pattern Recognition}, pp. 19358--19369. Vancouver, 2023.

\bibitem{ghiasi2021simple}
G.~Ghiasi, Y.~Cui, A.~Srinivas, R.~Qian, T.-Y. Lin, E.~D. Cubuk, Q.~V. Le, and
  B.~Zoph.
\newblock Simple copy-paste is a strong data augmentation method for instance
  segmentation.
\newblock In {\em Proceedings of the IEEE/CVF Conference on Computer Vision and
  Pattern Recognition}, pp. 2918--2928, 2021.
  \href{https://doi.org/10.1109/cvpr46437.2021.00294}
{doi: {{%
10\hspace{.1pt}\discretionary{.}{%
}{.}\hspace{.4pt}1109\discretionary{/}{%
}{/}cvpr46437\hspace{.1pt}\discretionary{.}{%
}{.}\hspace{.4pt}2021\hspace{.1pt}\discretionary{.}{%
}{.}\hspace{.4pt}00294}}}


\bibitem{girshick2015fast}
R.~Girshick.
\newblock Fast {R-CNN}.
\newblock In {\em Proceedings of the IEEE International Conference on Computer
  Vision}, pp. 1440--1448. Santiago, 2015.
  \href{https://doi.org/10.1109/ICCV.2015.169}
{doi: {{%
10\hspace{.1pt}\discretionary{.}{%
}{.}\hspace{.4pt}1109\discretionary{/}{%
}{/}ICCV\hspace{.1pt}\discretionary{.}{%
}{.}\hspace{.4pt}2015\hspace{.1pt}\discretionary{.}{%
}{.}\hspace{.4pt}169}}}


\bibitem{gleicher2020boxer}
M.~Gleicher, A.~Barve, X.~Yu, and F.~Heimerl.
\newblock Boxer: Interactive comparison of classifier results.
\newblock {\em Computer Graphics Forum}, 39(3):181--193, 2020.
  \href{https://doi.org/10.1111/cgf.13972}
{doi: {{%
10\hspace{.1pt}\discretionary{.}{%
}{.}\hspace{.4pt}1111\discretionary{/}{%
}{/}cgf\hspace{.1pt}\discretionary{.}{%
}{.}\hspace{.4pt}13972}}}


\bibitem{gortler2022neo}
J.~G{\"o}rtler, F.~Hohman, D.~Moritz, K.~Wongsuphasawat, D.~Ren, R.~Nair,
  M.~Kirchner, and K.~Patel.
\newblock Neo: Generalizing confusion matrix visualization to hierarchical and
  multi-output labels.
\newblock In {\em Proceedings of the ACM CHI Conference on Human Factors in
  Computing Systems}, pp. 1--13. New Orleans, 2022.
  \href{https://doi.org/10.1145/3491102.3501823}
{doi: {{%
10\hspace{.1pt}\discretionary{.}{%
}{.}\hspace{.4pt}1145\discretionary{/}{%
}{/}3491102\hspace{.1pt}\discretionary{.}{%
}{.}\hspace{.4pt}3501823}}}


\bibitem{gou2020vatld}
L.~Gou, L.~Zou, N.~Li, M.~Hofmann, A.~K. Shekar, A.~Wendt, and L.~Ren.
\newblock {VATLD}: A visual analytics system to assess, understand and improve
  traffic light detection.
\newblock {\em IEEE Transactions on Visualization and Computer Graphics},
  27(2):261--271, 2021. \href{https://doi.org/10.1109/TVCG.2020.3030350}
{doi: {{%
10\hspace{.1pt}\discretionary{.}{%
}{.}\hspace{.4pt}1109\discretionary{/}{%
}{/}TVCG\hspace{.1pt}\discretionary{.}{%
}{.}\hspace{.4pt}2020\hspace{.1pt}\discretionary{.}{%
}{.}\hspace{.4pt}3030350}}}


\bibitem{gratzl2013lineup}
S.~Gratzl, A.~Lex, N.~Gehlenborg, H.~Pfister, and M.~Streit.
\newblock {LineUp}: Visual analysis of multi-attribute rankings.
\newblock {\em IEEE Transactions on Visualization and Computer Graphics},
  19(12):2277--2286, 2013. \href{https://doi.org/10.1109/TVCG.2013.173}
{doi: {{%
10\hspace{.1pt}\discretionary{.}{%
}{.}\hspace{.4pt}1109\discretionary{/}{%
}{/}TVCG\hspace{.1pt}\discretionary{.}{%
}{.}\hspace{.4pt}2013\hspace{.1pt}\discretionary{.}{%
}{.}\hspace{.4pt}173}}}


\bibitem{han2007frequent}
J.~Han, H.~Cheng, D.~Xin, and X.~Yan.
\newblock Frequent pattern mining: current status and future directions.
\newblock {\em Data Mining and Knowledge Discovery}, 15(1):55--86, 2007.
  \href{https://doi.org/10.1007/s10618-006-0059-1}
{doi: {{%
10\hspace{.1pt}\discretionary{.}{%
}{.}\hspace{.4pt}1007\discretionary{/}{%
}{/}s10618\discretionary{%
}{-}{-}006\discretionary{%
}{-}{-}0059\discretionary{%
}{-}{-}1}}}


\bibitem{he2016deep}
K.~He, X.~Zhang, S.~Ren, and J.~Sun.
\newblock Deep residual learning for image recognition.
\newblock In {\em Proceedings of the IEEE/CVF Conference on Computer Vision and
  Pattern Recognition}, pp. 770--778. Las Vegas, 2016.
  \href{https://doi.org/10.1109/CVPR.2016.90}
{doi: {{%
10\hspace{.1pt}\discretionary{.}{%
}{.}\hspace{.4pt}1109\discretionary{/}{%
}{/}CVPR\hspace{.1pt}\discretionary{.}{%
}{.}\hspace{.4pt}2016\hspace{.1pt}\discretionary{.}{%
}{.}\hspace{.4pt}90}}}


\bibitem{he2021can}
W.~He, L.~Zou, A.~K. Shekar, L.~Gou, and L.~Ren.
\newblock Where can we help? a visual analytics approach to diagnosing and
  improving semantic segmentation of movable objects.
\newblock {\em IEEE Transactions on Visualization and Computer Graphics},
  28(1):1040--1050, 2021. \href{https://doi.org/10.1109/TVCG.2021.3114855}
{doi: {{%
10\hspace{.1pt}\discretionary{.}{%
}{.}\hspace{.4pt}1109\discretionary{/}{%
}{/}TVCG\hspace{.1pt}\discretionary{.}{%
}{.}\hspace{.4pt}2021\hspace{.1pt}\discretionary{.}{%
}{.}\hspace{.4pt}3114855}}}


\bibitem{heyen2020clavis}
F.~Heyen, T.~Munz, M.~Neumann, D.~Ortega, N.~T. Vu, D.~Weiskopf, and
  M.~Sedlmair.
\newblock {ClaVis}: An interactive visual comparison system for classifiers.
\newblock In {\em Proceedings of the ACM International Conference on Advanced
  Visual Interfaces}, pp. 1--9. Island of Ischia, 2020.
  \href{https://doi.org/10.1145/3399715.3399814}
{doi: {{%
10\hspace{.1pt}\discretionary{.}{%
}{.}\hspace{.4pt}1145\discretionary{/}{%
}{/}3399715\hspace{.1pt}\discretionary{.}{%
}{.}\hspace{.4pt}3399814}}}


\bibitem{hinterreiter2020confusionflow}
A.~Hinterreiter, P.~Ruch, H.~Stitz, M.~Ennemoser, J.~Bernard, H.~Strobelt, and
  M.~Streit.
\newblock {ConfusionFlow}: A model-agnostic visualization for temporal analysis
  of classifier confusion.
\newblock {\em IEEE Transactions on Visualization and Computer Graphics},
  28(2):1222--1236, 2022. \href{https://doi.org/10.1109/TVCG.2020.3012063}
{doi: {{%
10\hspace{.1pt}\discretionary{.}{%
}{.}\hspace{.4pt}1109\discretionary{/}{%
}{/}TVCG\hspace{.1pt}\discretionary{.}{%
}{.}\hspace{.4pt}2020\hspace{.1pt}\discretionary{.}{%
}{.}\hspace{.4pt}3012063}}}


\bibitem{hohman2018visual}
F.~Hohman, M.~Kahng, R.~Pienta, and D.~H. Chau.
\newblock Visual analytics in deep learning: An interrogative survey for the
  next frontiers.
\newblock {\em IEEE Transactions on Visualization and Computer Graphics},
  25(8):2674--2693, 2019. \href{https://doi.org/10.1109/TVCG.2018.2843369}
{doi: {{%
10\hspace{.1pt}\discretionary{.}{%
}{.}\hspace{.4pt}1109\discretionary{/}{%
}{/}TVCG\hspace{.1pt}\discretionary{.}{%
}{.}\hspace{.4pt}2018\hspace{.1pt}\discretionary{.}{%
}{.}\hspace{.4pt}2843369}}}


\bibitem{hoiem2012diagnosing}
D.~Hoiem, Y.~Chodpathumwan, and Q.~Dai.
\newblock Diagnosing error in object detectors.
\newblock In {\em Proceedings of the European Conference on Computer Vision},
  pp. 340--353. Florence, 2012.
  \href{https://doi.org/10.1007/978-3-642-33712-3_25}
{doi: {{%
10\hspace{.1pt}\discretionary{.}{%
}{.}\hspace{.4pt}1007\discretionary{/}{%
}{/}978\discretionary{%
}{-}{-}3\discretionary{%
}{-}{-}642\discretionary{%
}{-}{-}33712\discretionary{%
}{-}{-}3\_25}}}


\bibitem{hou2022shape}
L.~Hou, K.~Lu, J.~Xue, and Y.~Li.
\newblock Shape-adaptive selection and measurement for oriented object
  detection.
\newblock In {\em Proceedings of the AAAI Conference on Artificial
  Intelligence}, pp. 923--932, 2022.
  \href{https://doi.org/10.1609/aaai.v36i1.19975}
{doi: {{%
10\hspace{.1pt}\discretionary{.}{%
}{.}\hspace{.4pt}1609\discretionary{/}{%
}{/}aaai\hspace{.1pt}\discretionary{.}{%
}{.}\hspace{.4pt}v36i1\hspace{.1pt}\discretionary{.}{%
}{.}\hspace{.4pt}19975}}}


\bibitem{jadhav2009parametric}
D.~Jadhav and T.~Ramanathan.
\newblock Parametric and non-parametric estimation of value-at-risk.
\newblock {\em The Journal of Risk Model Validation}, 3(1):51, 2009.
  \href{https://doi.org/doi.org/10.5539/ijbm.v8n11p103}
{doi: {{%
doi\hspace{.1pt}\discretionary{.}{%
}{.}\hspace{.4pt}org\discretionary{/}{%
}{/}10\hspace{.1pt}\discretionary{.}{%
}{.}\hspace{.4pt}5539\discretionary{/}{%
}{/}ijbm\hspace{.1pt}\discretionary{.}{%
}{.}\hspace{.4pt}v8n11p103}}}


\bibitem{johnson2019survey}
J.~M. Johnson and T.~M. Khoshgoftaar.
\newblock Survey on deep learning with class imbalance.
\newblock {\em Journal of Big Data}, 6(1):1--54, 2019.
  \href{https://doi.org/10.1186/s40537-019-0192-5}
{doi: {{%
10\hspace{.1pt}\discretionary{.}{%
}{.}\hspace{.4pt}1186\discretionary{/}{%
}{/}s40537\discretionary{%
}{-}{-}019\discretionary{%
}{-}{-}0192\discretionary{%
}{-}{-}5}}}


\bibitem{kervadec2021boundary}
H.~Kervadec, J.~Bouchtiba, C.~Desrosiers, E.~Granger, J.~Dolz, and I.~B. Ayed.
\newblock Boundary loss for highly unbalanced segmentation.
\newblock {\em Medical Image Analysis}, 67:101851, 2021.
  \href{https://doi.org/10.1016/j.media.2020.101851}
{doi: {{%
10\hspace{.1pt}\discretionary{.}{%
}{.}\hspace{.4pt}1016\discretionary{/}{%
}{/}j\hspace{.1pt}\discretionary{.}{%
}{.}\hspace{.4pt}media\hspace{.1pt}\discretionary{.}{%
}{.}\hspace{.4pt}2020\hspace{.1pt}\discretionary{.}{%
}{.}\hspace{.4pt}101851}}}


\bibitem{lei2020geometric}
N.~Lei, D.~An, Y.~Guo, K.~Su, S.~Liu, Z.~Luo, S.-T. Yau, and X.~Gu.
\newblock A geometric understanding of deep learning.
\newblock {\em Engineering}, 6(3):361--374, 2020.
  \href{https://doi.org/10.1016/j.eng.2019.09.010}
{doi: {{%
10\hspace{.1pt}\discretionary{.}{%
}{.}\hspace{.4pt}1016\discretionary{/}{%
}{/}j\hspace{.1pt}\discretionary{.}{%
}{.}\hspace{.4pt}eng\hspace{.1pt}\discretionary{.}{%
}{.}\hspace{.4pt}2019\hspace{.1pt}\discretionary{.}{%
}{.}\hspace{.4pt}09\hspace{.1pt}\discretionary{.}{%
}{.}\hspace{.4pt}010}}}


\bibitem{li2022unified}
Z.~Li, X.~Wang, W.~Yang, J.~Wu, Z.~Zhang, Z.~Liu, M.~Sun, H.~Zhang, and S.~Liu.
\newblock A unified understanding of deep {NLP} models for text classification.
\newblock {\em IEEE Transactions on Visualization and Computer Graphics},
  28(12):4980--4994, 2022. \href{https://doi.org/10.1109/TVCG.2022.3184186}
{doi: {{%
10\hspace{.1pt}\discretionary{.}{%
}{.}\hspace{.4pt}1109\discretionary{/}{%
}{/}TVCG\hspace{.1pt}\discretionary{.}{%
}{.}\hspace{.4pt}2022\hspace{.1pt}\discretionary{.}{%
}{.}\hspace{.4pt}3184186}}}


\bibitem{lin2014microsoft}
T.-Y. Lin, M.~Maire, S.~Belongie, J.~Hays, P.~Perona, D.~Ramanan,
  P.~Doll{\'a}r, and C.~L. Zitnick.
\newblock Microsoft {COCO}: Common objects in context.
\newblock In {\em Proceedings of the European Conference on Computer Vision},
  pp. 740--755. Zürich, 2014.
  \href{https://doi.org/10.1007/978-3-319-10602-1_48}
{doi: {{%
10\hspace{.1pt}\discretionary{.}{%
}{.}\hspace{.4pt}1007\discretionary{/}{%
}{/}978\discretionary{%
}{-}{-}3\discretionary{%
}{-}{-}319\discretionary{%
}{-}{-}10602\discretionary{%
}{-}{-}1\_48}}}


\bibitem{liu2020deep}
L.~Liu, W.~Ouyang, X.~Wang, P.~Fieguth, J.~Chen, X.~Liu, and
  M.~Pietik{\"a}inen.
\newblock Deep learning for generic object detection: A survey.
\newblock {\em International Journal of Computer Vision}, 128:261--318, 2020.
  \href{https://doi.org/10.1007/s11263-019-01247-4}
{doi: {{%
10\hspace{.1pt}\discretionary{.}{%
}{.}\hspace{.4pt}1007\discretionary{/}{%
}{/}s11263\discretionary{%
}{-}{-}019\discretionary{%
}{-}{-}01247\discretionary{%
}{-}{-}4}}}


\bibitem{liu2018analyzing}
M.~Liu, S.~Liu, H.~Su, K.~Cao, and J.~Zhu.
\newblock Analyzing the noise robustness of deep neural networks.
\newblock In {\em Proceedings of the IEEE Conference on Visual Analytics
  Science and Technology}, pp. 60--71. Berlin, 2018.
  \href{https://doi.org/10.1109/VAST.2018.8802509}
{doi: {{%
10\hspace{.1pt}\discretionary{.}{%
}{.}\hspace{.4pt}1109\discretionary{/}{%
}{/}VAST\hspace{.1pt}\discretionary{.}{%
}{.}\hspace{.4pt}2018\hspace{.1pt}\discretionary{.}{%
}{.}\hspace{.4pt}8802509}}}


\bibitem{liu2016towards}
M.~Liu, J.~Shi, Z.~Li, C.~Li, J.~Zhu, and S.~Liu.
\newblock Towards better analysis of deep convolutional neural networks.
\newblock {\em IEEE Transactions on Visualization and Computer Graphics},
  23(1):91--100, 2017. \href{https://doi.org/10.1109/TVCG.2016.2598831}
{doi: {{%
10\hspace{.1pt}\discretionary{.}{%
}{.}\hspace{.4pt}1109\discretionary{/}{%
}{/}TVCG\hspace{.1pt}\discretionary{.}{%
}{.}\hspace{.4pt}2016\hspace{.1pt}\discretionary{.}{%
}{.}\hspace{.4pt}2598831}}}


\bibitem{liu2021catnet}
Y.~Liu, H.~Li, C.~Hu, S.~Luo, H.~Shen, and C.~W. Chen.
\newblock Learning to aggregate multi-scale context for instance segmentation
  in remote sensing images.
\newblock {\em CoRR}, abs/2111.11057, 2021.
  \href{https://doi.org/10.48550/arXiv.2111.11057}
{doi: {{%
10\hspace{.1pt}\discretionary{.}{%
}{.}\hspace{.4pt}48550\discretionary{/}{%
}{/}arXiv\hspace{.1pt}\discretionary{.}{%
}{.}\hspace{.4pt}2111\hspace{.1pt}\discretionary{.}{%
}{.}\hspace{.4pt}11057}}}


\bibitem{liu2021swin}
Z.~Liu, Y.~Lin, Y.~Cao, H.~Hu, Y.~Wei, Z.~Zhang, S.~Lin, and B.~Guo.
\newblock Swin transformer: Hierarchical vision transformer using shifted
  windows.
\newblock In {\em Proceedings of the IEEE/CVF International Conference on
  Computer Vision}, pp. 10012--10022. Montreal, 2021.
  \href{https://doi.org/10.1109/iccv48922.2021.00986}
{doi: {{%
10\hspace{.1pt}\discretionary{.}{%
}{.}\hspace{.4pt}1109\discretionary{/}{%
}{/}iccv48922\hspace{.1pt}\discretionary{.}{%
}{.}\hspace{.4pt}2021\hspace{.1pt}\discretionary{.}{%
}{.}\hspace{.4pt}00986}}}


\bibitem{liu2022convnet}
Z.~Liu, H.~Mao, C.-Y. Wu, C.~Feichtenhofer, T.~Darrell, and S.~Xie.
\newblock A {ConvNet} for the 2020s.
\newblock In {\em Proceedings of the IEEE/CVF Conference on Computer Vision and
  Pattern Recognition}, pp. 11976--11986. New Orleans, 2022.
  \href{https://doi.org/10.1109/cvpr52688.2022.01167}
{doi: {{%
10\hspace{.1pt}\discretionary{.}{%
}{.}\hspace{.4pt}1109\discretionary{/}{%
}{/}cvpr52688\hspace{.1pt}\discretionary{.}{%
}{.}\hspace{.4pt}2022\hspace{.1pt}\discretionary{.}{%
}{.}\hspace{.4pt}01167}}}


\bibitem{maaten2008visualizing}
L.~v.~d. Maaten and G.~Hinton.
\newblock Visualizing data using t-{SNE}.
\newblock {\em Journal of Machine Learning Research}, 9(86):2579--2605, 2008.

\bibitem{minaee2021image}
S.~Minaee, Y.~Y. Boykov, F.~Porikli, A.~J. Plaza, N.~Kehtarnavaz, and
  D.~Terzopoulos.
\newblock Image segmentation using deep learning: A survey.
\newblock {\em IEEE Transactions on Pattern Analysis and Machine Intelligence},
  44(7):3523--3542, 2022. \href{https://doi.org/10.1109/TPAMI.2021.3059968}
{doi: {{%
10\hspace{.1pt}\discretionary{.}{%
}{.}\hspace{.4pt}1109\discretionary{/}{%
}{/}TPAMI\hspace{.1pt}\discretionary{.}{%
}{.}\hspace{.4pt}2021\hspace{.1pt}\discretionary{.}{%
}{.}\hspace{.4pt}3059968}}}


\bibitem{murtagh2012algorithms}
F.~Murtagh and P.~Contreras.
\newblock Algorithms for hierarchical clustering: an overview.
\newblock {\em Wiley Interdisciplinary Reviews: Data Mining and Knowledge
  Discovery}, 2(1):86--97, 2012. \href{https://doi.org/10.1002/widm.53}
{doi: {{%
10\hspace{.1pt}\discretionary{.}{%
}{.}\hspace{.4pt}1002\discretionary{/}{%
}{/}widm\hspace{.1pt}\discretionary{.}{%
}{.}\hspace{.4pt}53}}}


\bibitem{nakkiran2021deep}
P.~Nakkiran, G.~Kaplun, Y.~Bansal, T.~Yang, B.~Barak, and I.~Sutskever.
\newblock Deep double descent: Where bigger models and more data hurt.
\newblock {\em Journal of Statistical Mechanics: Theory and Experiment},
  2021(12):124003, 2021. \href{https://doi.org/10.1088/1742-5468/ac3a74}
{doi: {{%
10\hspace{.1pt}\discretionary{.}{%
}{.}\hspace{.4pt}1088\discretionary{/}{%
}{/}1742\discretionary{%
}{-}{-}5468\discretionary{/}{%
}{/}ac3a74}}}


\bibitem{pastor2021looking}
E.~Pastor, L.~de~Alfaro, and E.~Baralis.
\newblock Looking for trouble: Analyzing classifier behavior via pattern
  divergence.
\newblock In {\em Proceedings of the ACM International Conference on Management
  of Data}, pp. 1400--1412. Xi'an, 2021.
  \href{https://doi.org/10.1145/3448016.3457284}
{doi: {{%
10\hspace{.1pt}\discretionary{.}{%
}{.}\hspace{.4pt}1145\discretionary{/}{%
}{/}3448016\hspace{.1pt}\discretionary{.}{%
}{.}\hspace{.4pt}3457284}}}


\bibitem{pentico2007assignment}
D.~W. Pentico.
\newblock Assignment problems: A golden anniversary survey.
\newblock {\em European Journal of Operational Research}, 176(2):774--793,
  2007. \href{https://doi.org/10.1016/j.ejor.2005.09.014}
{doi: {{%
10\hspace{.1pt}\discretionary{.}{%
}{.}\hspace{.4pt}1016\discretionary{/}{%
}{/}j\hspace{.1pt}\discretionary{.}{%
}{.}\hspace{.4pt}ejor\hspace{.1pt}\discretionary{.}{%
}{.}\hspace{.4pt}2005\hspace{.1pt}\discretionary{.}{%
}{.}\hspace{.4pt}09\hspace{.1pt}\discretionary{.}{%
}{.}\hspace{.4pt}014}}}


\bibitem{ren2016squares}
D.~Ren, S.~Amershi, B.~Lee, J.~Suh, and J.~D. Williams.
\newblock Squares: Supporting interactive performance analysis for multiclass
  classifiers.
\newblock {\em IEEE Transactions on Visualization and Computer Graphics},
  23(1):61--70, 2017. \href{https://doi.org/10.1109/TVCG.2016.2598828}
{doi: {{%
10\hspace{.1pt}\discretionary{.}{%
}{.}\hspace{.4pt}1109\discretionary{/}{%
}{/}TVCG\hspace{.1pt}\discretionary{.}{%
}{.}\hspace{.4pt}2016\hspace{.1pt}\discretionary{.}{%
}{.}\hspace{.4pt}2598828}}}


\bibitem{rottmann2023mosaicsets}
P.~Rottmann, M.~Wallinger, A.~Bonerath, S.~Gedicke, M.~Nöllenburg, and J.-H.
  Haunert.
\newblock {MosaicSets}: Embedding set systems into grid graphs.
\newblock {\em IEEE Transactions on Visualization and Computer Graphics},
  29(1):875--885, 2023. \href{https://doi.org/10.1109/TVCG.2022.3209485}
{doi: {{%
10\hspace{.1pt}\discretionary{.}{%
}{.}\hspace{.4pt}1109\discretionary{/}{%
}{/}TVCG\hspace{.1pt}\discretionary{.}{%
}{.}\hspace{.4pt}2022\hspace{.1pt}\discretionary{.}{%
}{.}\hspace{.4pt}3209485}}}


\bibitem{schroeder2008switching}
W.~Schroeder.
\newblock Switching between tools in complex applications.
\newblock {\em Journal of Usability Studies}, 3(4):173--188, 2008.

\bibitem{sevastjanova2022visual}
R.~Sevastjanova, E.~Cakmak, S.~Ravfogel, R.~Cotterell, and M.~El-Assady.
\newblock Visual comparison of language model adaptation.
\newblock {\em IEEE Transactions on Visualization and Computer Graphics},
  29(1):1178--1188, 2023. \href{https://doi.org/10.1109/TVCG.2022.3209458}
{doi: {{%
10\hspace{.1pt}\discretionary{.}{%
}{.}\hspace{.4pt}1109\discretionary{/}{%
}{/}TVCG\hspace{.1pt}\discretionary{.}{%
}{.}\hspace{.4pt}2022\hspace{.1pt}\discretionary{.}{%
}{.}\hspace{.4pt}3209458}}}


\bibitem{townsend1971theoretical}
J.~T. Townsend.
\newblock Theoretical analysis of an alphabetic confusion matrix.
\newblock {\em Perception \& Psychophysics}, 9:40--50, 1971.
  \href{https://doi.org/10.3758/BF03213026}
{doi: {{%
10\hspace{.1pt}\discretionary{.}{%
}{.}\hspace{.4pt}3758\discretionary{/}{%
}{/}BF03213026}}}


\bibitem{wang2023drava}
Q.~Wang, S.~L'Yi, and N.~Gehlenborg.
\newblock {DRAVA}: Aligning human concepts with machine learning latent
  dimensions for the visual exploration of small multiples.
\newblock In {\em Proceedings of the ACM CHI Conference on Human Factors in
  Computing Systems}, pp. 1--15. Hamburg, 2023.
  \href{https://doi.org/10.1145/3544548.3581127}
{doi: {{%
10\hspace{.1pt}\discretionary{.}{%
}{.}\hspace{.4pt}1145\discretionary{/}{%
}{/}3544548\hspace{.1pt}\discretionary{.}{%
}{.}\hspace{.4pt}3581127}}}


\bibitem{wang2022internimage}
W.~Wang, J.~Dai, Z.~Chen, Z.~Huang, Z.~Li, X.~Zhu, X.~Hu, T.~Lu, L.~Lu, H.~Li,
  et~al.
\newblock {InternImage}: Exploring large-scale vision foundation models with
  deformable convolutions.
\newblock In {\em Proceedings of the IEEE/CVF Conference on Computer Vision and
  Pattern Recognition}, pp. 14408--14419. Vancouver, 2023.

\bibitem{waqas2019isaid}
S.~Waqas~Zamir, A.~Arora, A.~Gupta, S.~Khan, G.~Sun, F.~Shahbaz~Khan, F.~Zhu,
  L.~Shao, G.-S. Xia, and X.~Bai.
\newblock {iSAID}: A large-scale dataset for instance segmentation in aerial
  images.
\newblock In {\em Proceedings of the IEEE/CVF Conference on Computer Vision and
  Pattern Recognition Workshops}, pp. 28--37. Long Beach, 2019.

\bibitem{willett2007scented}
W.~Willett, J.~Heer, and M.~Agrawala.
\newblock Scented widgets: Improving navigation cues with embedded
  visualizations.
\newblock {\em IEEE Transactions on Visualization and Computer Graphics},
  13(6):1129--1136, 2007. \href{https://doi.org/10.1109/TVCG.2007.70589}
{doi: {{%
10\hspace{.1pt}\discretionary{.}{%
}{.}\hspace{.4pt}1109\discretionary{/}{%
}{/}TVCG\hspace{.1pt}\discretionary{.}{%
}{.}\hspace{.4pt}2007\hspace{.1pt}\discretionary{.}{%
}{.}\hspace{.4pt}70589}}}


\bibitem{wong1987synthesizing}
A.~K. Wong and D.~K. Chiu.
\newblock Synthesizing statistical knowledge from incomplete mixed-mode data.
\newblock {\em IEEE Transactions on Pattern Analysis and Machine Intelligence},
  9(6):796--805, 1987. \href{https://doi.org/10.1109/TPAMI.1987.4767986}
{doi: {{%
10\hspace{.1pt}\discretionary{.}{%
}{.}\hspace{.4pt}1109\discretionary{/}{%
}{/}TPAMI\hspace{.1pt}\discretionary{.}{%
}{.}\hspace{.4pt}1987\hspace{.1pt}\discretionary{.}{%
}{.}\hspace{.4pt}4767986}}}


\bibitem{xenopoulos2022calibrate}
P.~Xenopoulos, J.~Rulff, L.~G. Nonato, B.~Barr, and C.~Silva.
\newblock Calibrate: Interactive analysis of probabilistic model output.
\newblock {\em IEEE Transactions on Visualization and Computer Graphics},
  29(1):853--863, 2023. \href{https://doi.org/10.1109/TVCG.2022.3209489}
{doi: {{%
10\hspace{.1pt}\discretionary{.}{%
}{.}\hspace{.4pt}1109\discretionary{/}{%
}{/}TVCG\hspace{.1pt}\discretionary{.}{%
}{.}\hspace{.4pt}2022\hspace{.1pt}\discretionary{.}{%
}{.}\hspace{.4pt}3209489}}}


\bibitem{yang2022focal}
J.~Yang, C.~Li, X.~Dai, and J.~Gao.
\newblock Focal modulation networks.
\newblock In {\em Proceedings of the Advances in Neural Information Processing
  Systems}, pp. 4203--4217. New Orleans, 2022.
  \href{https://doi.org/10.31525/ct1-nct03946618}
{doi: {{%
10\hspace{.1pt}\discretionary{.}{%
}{.}\hspace{.4pt}31525\discretionary{/}{%
}{/}ct1\discretionary{%
}{-}{-}nct03946618}}}


\bibitem{yang2020drift}
W.~{Yang}, Z.~{Li}, M.~{Liu}, Y.~{Lu}, K.~{Cao}, R.~{Maciejewski}, and
  S.~{Liu}.
\newblock Diagnosing concept drift with visual analytics.
\newblock In {\em Proceedings of IEEE Conference on Visual Analytics Science
  and Technology}, pp. 12--23, 2020.
  \href{https://doi.org/10.1109/vast50239.2020.00007}
{doi: {{%
10\hspace{.1pt}\discretionary{.}{%
}{.}\hspace{.4pt}1109\discretionary{/}{%
}{/}vast50239\hspace{.1pt}\discretionary{.}{%
}{.}\hspace{.4pt}2020\hspace{.1pt}\discretionary{.}{%
}{.}\hspace{.4pt}00007}}}


\bibitem{yang2021interactive}
W.~Yang, X.~Wang, J.~Lu, W.~Dou, and S.~Liu.
\newblock Interactive steering of hierarchical clustering.
\newblock {\em IEEE Transactions on Visualization and Computer Graphics},
  27(10):3953--3967, 2021. \href{https://doi.org/10.1109/tvcg.2020.2995100}
{doi: {{%
10\hspace{.1pt}\discretionary{.}{%
}{.}\hspace{.4pt}1109\discretionary{/}{%
}{/}tvcg\hspace{.1pt}\discretionary{.}{%
}{.}\hspace{.4pt}2020\hspace{.1pt}\discretionary{.}{%
}{.}\hspace{.4pt}2995100}}}


\bibitem{yang2022diagnosing}
W.~Yang, X.~Ye, X.~Zhang, L.~Xiao, J.~Xia, Z.~Wang, J.~Zhu, H.~Pfister, and
  S.~Liu.
\newblock Diagnosing ensemble few-shot classifiers.
\newblock {\em IEEE Transactions on Visualization and Computer Graphics},
  28(9):3292--3306, 2022. \href{https://doi.org/10.1109/TVCG.2022.3182488}
{doi: {{%
10\hspace{.1pt}\discretionary{.}{%
}{.}\hspace{.4pt}1109\discretionary{/}{%
}{/}TVCG\hspace{.1pt}\discretionary{.}{%
}{.}\hspace{.4pt}2022\hspace{.1pt}\discretionary{.}{%
}{.}\hspace{.4pt}3182488}}}


\bibitem{yuan2021survey}
J.~Yuan, C.~Chen, W.~Yang, M.~Liu, J.~Xia, and S.~Liu.
\newblock A survey of visual analytics techniques for machine learning.
\newblock {\em Computational Visual Media}, 7(1):3--36, 2021.
  \href{https://doi.org/10.1007/s41095-020-0191-7}
{doi: {{%
10\hspace{.1pt}\discretionary{.}{%
}{.}\hspace{.4pt}1007\discretionary{/}{%
}{/}s41095\discretionary{%
}{-}{-}020\discretionary{%
}{-}{-}0191\discretionary{%
}{-}{-}7}}}


\bibitem{zhang2022dino}
H.~Zhang, F.~Li, S.~Liu, L.~Zhang, H.~Su, J.~Zhu, L.~Ni, and H.~Shum.
\newblock {DINO}: {DETR} with improved denoising anchor boxes for end-to-end
  object detection.
\newblock In {\em Proceedings of the International Conference on Learning
  Representations}, pp. 1--9. Kigali, 2023.

\bibitem{zhang2022sliceteller}
X.~Zhang, J.~P. Ono, H.~Song, L.~Gou, K.-L. Ma, and L.~Ren.
\newblock {SliceTeller}: A data slice-driven approach for machine learning
  model validation.
\newblock {\em IEEE Transactions on Visualization and Computer Graphics},
  29(1):842--852, 2023. \href{https://doi.org/10.1109/TVCG.2022.3209465}
{doi: {{%
10\hspace{.1pt}\discretionary{.}{%
}{.}\hspace{.4pt}1109\discretionary{/}{%
}{/}TVCG\hspace{.1pt}\discretionary{.}{%
}{.}\hspace{.4pt}2022\hspace{.1pt}\discretionary{.}{%
}{.}\hspace{.4pt}3209465}}}


\bibitem{zhang2018overview}
Y.~Zhang and Q.~Yang.
\newblock An overview of multi-task learning.
\newblock {\em National Science Review}, 5(1):30--43, 2018.
  \href{https://doi.org/10.1093/nsr/nwx105}
{doi: {{%
10\hspace{.1pt}\discretionary{.}{%
}{.}\hspace{.4pt}1093\discretionary{/}{%
}{/}nsr\discretionary{/}{%
}{/}nwx105}}}


\bibitem{zhang2021survey}
Y.~Zhang and Q.~Yang.
\newblock A survey on multi-task learning.
\newblock {\em IEEE Transactions on Knowledge and Data Engineering},
  34(12):5586--5609, 2022. \href{https://doi.org/10.1109/TKDE.2021.3070203}
{doi: {{%
10\hspace{.1pt}\discretionary{.}{%
}{.}\hspace{.4pt}1109\discretionary{/}{%
}{/}TKDE\hspace{.1pt}\discretionary{.}{%
}{.}\hspace{.4pt}2021\hspace{.1pt}\discretionary{.}{%
}{.}\hspace{.4pt}3070203}}}


\end{thebibliography}

\includepdf[pages=-]{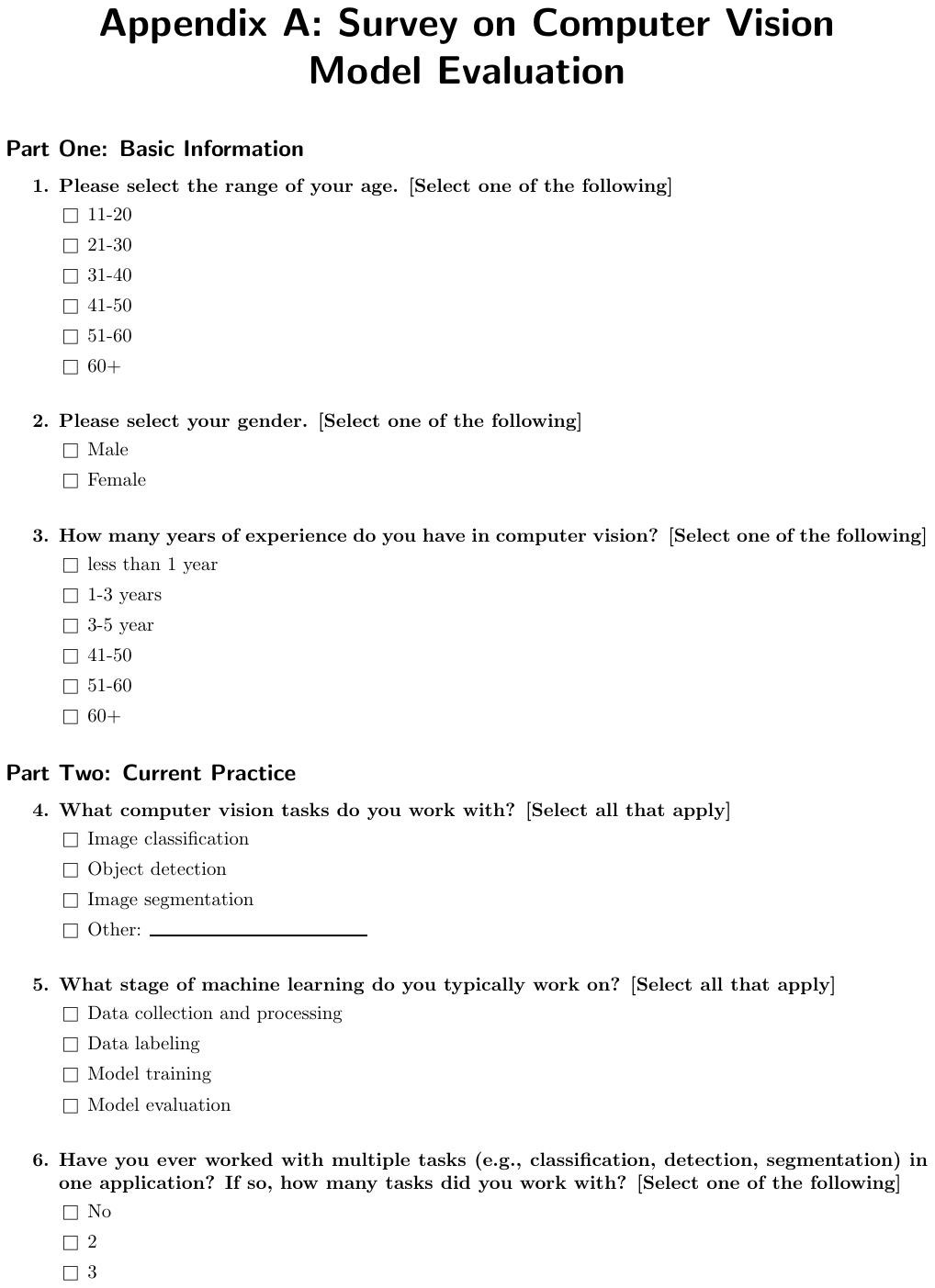}

\end{document}


\maketitle
\vspace{-22mm}


\subsection*{Part One: Basic Information}
\Qitem{ \Qq{Please select the range of your age. [Select one of the following]}
\begin{Qlist}
\item 11-20
\item 21-30
\item 31-40
\item 41-50
\item 51-60
\item 60+
\end{Qlist}
}

\Qitem{ \Qq{Please select your gender. [Select one of the following]}
\begin{Qlist}
\item Male
\item Female
\end{Qlist}
}

\thispagestyle{empty}

\Qitem{ \Qq{How many years of experience do you have in computer vision? [Select one of the following]}
\begin{Qlist}
\item less than 1 year
\item 1-3 years
\item 3-5 year
\item 41-50
\item 51-60
\item 60+
\end{Qlist}
}

\subsection*{Part Two: Current Practice}
\Qitem{ \Qq{What computer vision tasks do you work with? [Select all that apply] }
\begin{Qlist}
\item Image classification
\item Object detection
\item Image segmentation
\item Other: \Qline{4cm}
\end{Qlist}
}

\Qitem{ \Qq{What stage of machine learning do you typically work on? [Select all that apply] }
\begin{Qlist}
\item Data collection and processing
\item Data labeling
\item Model training
\item Model evaluation
\end{Qlist}
}

\Qitem{ \Qq{Have you ever worked with multiple tasks (e.g., classification, detection, segmentation) in one application? If so, how many tasks did you work with? [Select one of the following] }
\begin{Qlist}
\item No
\item 2
\item 3
\item 4
\item 4+
\end{Qlist}
}

\Qitem{ \Qq{How do you evaluate a computer vision model? [Select all that apply] }
\begin{Qlist}
\item By analyzing the training logs of the model, including \Qline{4cm}
\item By using visualization tools, including \Qline{4cm}
\item By inspecting the prediction results of a given sample
\item Other: \Qline{4cm}
\end{Qlist}
}

\Qitem{ \Qq{What problems do you encounter when using these evaluation methods? [Select all that apply] }
\begin{Qlist}
\item These methods focus on evaluating classification results and do not support the evaluation of more complex prediction results, such as detection results
\item Cannot help users identify problems in the training data, e.g., incorrect annotations
\item Cannot help users identify the classes where the model performs poorly
\item Cannot help users compare different models at the class level and instance level
\item Other: \Qline{4cm}
\end{Qlist}
}

\subsection*{Part Three: Key Features Needed}

\textbf{Do you think the following features can help you analyze and improve computer vision model performance? Please rate the importance.}

Unimportant / Slightly important / Moderately important / Important / Very important

\Qitem{ \Qq{A unified evaluation for different computer vision tasks.}

{Unimportant \Qrating{5} Very important}}

\Qitem{ \Qq{Analyzing the overall model performance on the entire dataset.}

{Unimportant \Qrating{5} Very important}}

\Qitem{ \Qq{Analyzing the model performance on data subsets (e.g., objects with large/small sizes).}

{Unimportant \Qrating{5} Very important}}

\Qitem{ \Qq{Exploring the prediction results of a given sample (e.g., a mispredicted sample) efficiently.}

{Unimportant \Qrating{5} Very important}}

\clearpage